\pdfoutput=1
\documentclass[11pt]{article}

\usepackage[]{EMNLP2023}
\usepackage{xcolor}
\usepackage{booktabs}
\usepackage{bbding}
\usepackage{fdsymbol}
\usepackage{tabularx}
\usepackage{multirow}
\usepackage{amsmath}
\usepackage{epigraph}
\usepackage{amsfonts}
\usepackage{pgfplots}
\pgfplotsset{compat=1.7}
\usepackage{subcaption}
\usepackage{pifont}
\usepackage{colortbl}
\usepackage{algorithmic}
\usepackage[ruled,vlined]{algorithm2e}
\usepackage{booktabs}
\usepackage{CJK}
\usepackage{times}
\usepackage{latexsym}
\usepackage{bbding}
\usepackage[T1]{fontenc}

\usepackage[utf8]{inputenc}
\usepackage{microtype}
\usepackage{inconsolata}

\newcommand{\turbo}{\texttt{GPT-3.5-Turbo-0301}\xspace}
\newcommand{\gpt}{\texttt{GPT-4-0314}\xspace} 
\newcommand{\claude}{\texttt{Claude-2}\xspace}
\newcommand{\vari}[1]{\small{$\pm$#1}}
\newcommand{\gptn}{\texttt{GPT-4}\xspace} 
\newcommand{\turbon}{\texttt{GPT-3.5}\xspace}

\title{Exchange-of-Thought: Enhancing Large Language Model Capabilities through Cross-Model Communication}

\author{
Zhangyue Yin\textsuperscript{$\diamondsuit$}\quad \quad
Qiushi Sun\textsuperscript{$\heartsuit$} \quad \quad
Cheng Chang\textsuperscript{$\diamondsuit$}\\
\bf{
Qipeng Guo\textsuperscript{$\diamondsuit\clubsuit$} \quad \quad
Junqi Dai\textsuperscript{$\diamondsuit$} \quad \quad
Xuanjing Huang\textsuperscript{$\diamondsuit$}\quad \quad
Xipeng Qiu\textsuperscript{$\diamondsuit$}\thanks{\ \ \ Corresponding author.}
}\\
\textsuperscript{$\diamondsuit$}School of Computer Science, Fudan University \\
\textsuperscript{$\heartsuit$}National University of Singapore 
\textsuperscript{$\clubsuit$}Shanghai AI Laboratory \\
\texttt{\{yinzy21,changc21,jqdai22\}@m.fudan.edu.cn} \quad \texttt{qiushisun@u.nus.edu} \\
\texttt{\{qpguo16, xjhuang, xpqiu\}@fudan.edu.cn}
}

\begin{document}
\maketitle

\begin{abstract}
Large Language Models (LLMs) have recently made significant strides in complex reasoning tasks through the Chain-of-Thought technique. 
Despite this progress, their reasoning is often constrained by their intrinsic understanding, lacking external insights. 
To address this, we propose Exchange-of-Thought (EoT), a novel framework that enables cross-model communication during problem-solving. 
Drawing inspiration from network topology, EoT integrates four unique communication paradigms: Memory, Report, Relay, and Debate.
This paper delves into the communication dynamics and volume associated with each paradigm.
To counterbalance the risks of incorrect reasoning chains, we implement a robust confidence evaluation mechanism within these communications.
Our experiments across diverse complex reasoning tasks demonstrate that EoT significantly surpasses established baselines, underscoring the value of external insights in enhancing LLM performance. 
Furthermore, we show that EoT achieves these superior results in a cost-effective manner, marking a promising advancement for efficient and collaborative AI problem-solving.

\end{abstract}
\vspace{-0.6em}

\vspace{0.2in}
\noindent ``Two heads are better than one.'' \\
\rightline{--\textit{English Proverb}}

\label{sec:introduction}

\section{Introduction}
Large Language Models (LLMs) such as GPT-4~\citep{openai2023gpt4} are revolutionizing the field of Natural Language Processing (NLP) by utilizing vast training corpora and huge computational resources~\citep[][\emph{inter alia}]{bai2022training, ouyang2022training, chowdhery2022palm, zhang2022opt, touvron2023llama}.
Although LLMs achieve exemplary performance across a wide range of NLP tasks~\citep{wei2022emergent, chung2022scaling}, they consistently struggle to perform well in reasoning tasks, and this limitation cannot be overcome solely by increasing the size of models~\citep{rae2021scaling, srivastava2023bb}.

To overcome this shortcoming, \citet{wei2022chain} proposed chain-of-thought (CoT) prompting, which guides the model to generate a series of intermediate reasoning steps before reaching the final answer. 
At the same time, 
a series of self-correction methods~\citep{welleck2023generating, ganguli2023capacity}
have been proposed,
which aim to iteratively improve the quality of answers by leveraging the model's feedback to their previous outputs~\citep{madaan2023self, shinn2023reflexion}.

\begin{figure}[t]
    \centering
 \includegraphics[width=\linewidth]{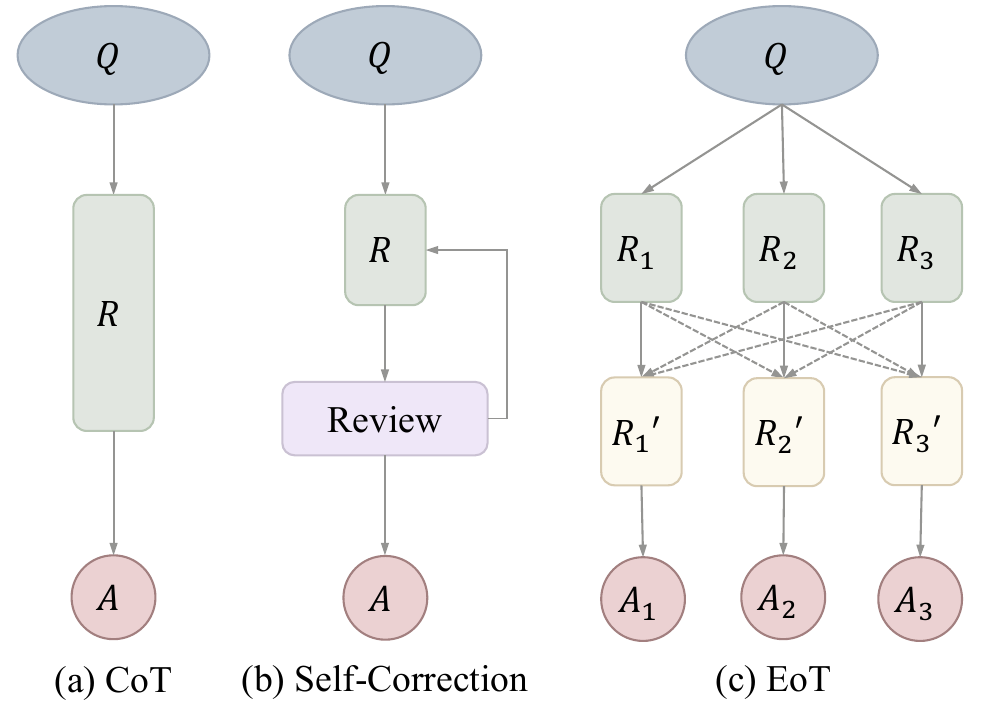}
    \caption{Comparison of CoT, Self-Correction, and EoT. Both CoT and Self-Correction rely on the model's innate abilities to generate and refine output, lacking external insights. EoT enhances the model's reasoning ability by incorporating the thoughts of other models as external insights.}
    \label{fig:intro}
    \vspace{-1em}
\end{figure}

However, CoT and self-correction solely base on the model's own understanding and perspective of the question during the reasoning process. Recent studies~\citep{huang2023large, valmeekam2023large, stechly2023gpt4} indicate that LLMs struggle to revise their responses without external feedback. This can be attributed to the model's complete dependence on internal representations to generate responses, which makes it difficult to overcome inherent limitations in capability~\citep{yin2023selfaware}.

\begin{figure}[t]
    \centering
 \includegraphics[width=\linewidth]{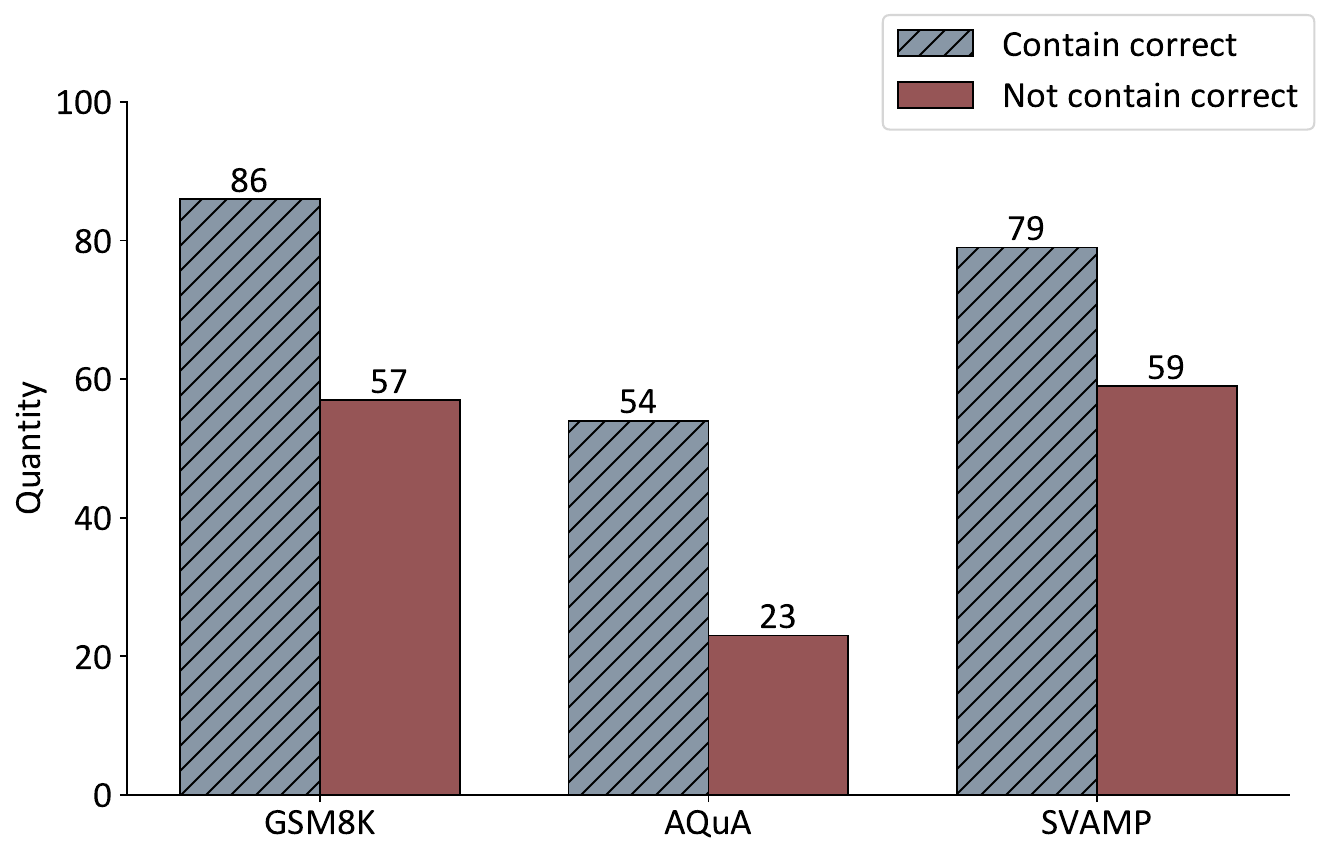}
    \caption{Pilot experiments on three reasoning datasets. The number of erroneous samples containing the correct answer is significantly higher than those not containing the correct answer.}
    \label{fig:prior-exp}
    \vspace{-1em}
\end{figure}

Despite the undeniable importance of external insights~\citep{yao2023react}, acquiring high-quality external insights remains a challenge. \citet{wang2023sc}'s research suggests that the single reasoning chain generated by CoT limits the model's reasoning performance. 
By increasing the temperature to sample diverse reasoning chains and selecting answers through majority voting, the model's reasoning performance can be further improved. 
However,
when confronted with difficult questions, 
the model often yields a higher number of incorrect responses.
In Figure~\ref{fig:prior-exp}, 
our analysis of correct and incorrect answers within erroneous samples from three reasoning datasets reveals that in most cases the model can deduce the correct answer.

In human society, the truth, even when held by a minority, can gain widespread acceptance and recognition through clear and persuasive communication~\citep{le1897crowd}.
The correct reasoning of others can serve as high-quality external insights, enriching and elevating our collective understanding.
Thus, 
we propose Exchange-of-Thought (EoT), 
a novel framework that fosters cross-model communication during the problem-solving process. 
This initiative enables models to incorporate the reasoning of others as external insights.

Figure~\ref{fig:intro} contrasts EoT with CoT and self-correction methods, highlighting the unique approach of EoT in integrating external perspectives. 
Inspired by the principles of network topology~\citep{bisht2015analytical} and agent communication~\citep{parsons2003argument}, 
we propose four communication paradigms: Memory, Report, Relay, and Debate. 
These paradigms are designed to facilitate the exchange of ideas and reasoning chains among models, enriching the problem-solving process with a diversity of insights.
Furthermore, we delve into the intricacies of each communication paradigm, 
analyzing the dynamics of information flow and the volume of communication. 
With the awareness that both correct and incorrect reasoning chains propagate within communications, 
we introduce confidence evaluation mechanisms that employs the analysis of answer variations to assess models' confidence levels.
It is designed to mitigate the influence of erroneous reasoning, thereby ensuring the integrity and reliability of the problem-solving process.

Experiments across various complex reasoning tasks demonstrate that EoT significantly outperforms established strong baselines, 
underscoring the critical role of external insights in augmenting the capabilities of LLMs.
We summarize our contributions as follows:
\vspace{-0.5em}
\begin{itemize}

\item We introduce Exchange-of-Thought (EoT), 
a pioneering framework for cross-model communication that incorporates external insights from other LLMs during problem-solving.
\vspace{-0.5em}
\item We present and examine four communication paradigms coupled with a confidence evaluation mechanism that assesses model certainty through the variability of answers, mitigating the impact of incorrect reasoning.
\vspace{-0.5em}
\item Experimental results on various complex reasoning tasks underscore the efficacy and cost-effectiveness of EoT, highlighting the significance of incorporating external insights and communication in problem-solving.
\end{itemize}

\section{Related Work}
\label{sec:related-work}
\subsection{Chain-of-Thought prompting in LLMs} 

\citet{wei2022chain} highlight that LLMs can manifest enhanced reasoning capabilities when being prompted by demonstrations with intermediate reasoning steps. 
This technique can effectively improve the performance of LLMs on complex reasoning tasks~\citep{wei2022emergent,kojima2022large}.
A series of strategies for enhancing CoT has been proposed to further improve the performance of LLMs. One such method is program-aided language models~\citep{gao2022pal, chen2022program}, 
which aims to decouple reasoning and computation through program synthesis. 
Moreover, complex tasks can also be transformed into delegable sub-tasks through modular approaches~\citep{khot2023decomp}.
Choosing appropriate demonstrations can also enhance the performance of CoT~\citep{li2023unified, li2023finding}. 
Notable among these,
AutoCoT~\citep{zhang2022automatic} uses an automated way to construct and sample diverse demonstrations. 
Active-Prompt~\citep{diao2023active} selects the most helpful samples for labeling based on the model's uncertainty in the outputs.
Recently,
\citet{li2023mot} employ a strategy of storing high-confidence thoughts as external memory and retrieves these insights to aid the reasoning process.

\subsection{Ensemble of Reasoning Paths}
LLMs have the ability to explore multiple reasoning paths using techniques such as temperature adjustment and prompt sampling~\citep{chuCoTReasoningSurvey2023}. \citet{wang2023sc} suggest that for complex questions, 
there may be several correct paths to approach a problem, 
leading to the proposal of Self-Consistency. 
This method replaces the greedy decoding strategy with the sampling of multiple reasoning paths and selecting the most consistent answer, resulting in significant performance improvements.
Beyond that,
\citet{fu2023complexcot} discover that prompts with higher reasoning complexity could achieve better performance in multi-step reasoning tasks, leading to the proposal of complexity-based prompting.
While other methods, 
such as re-ranking~\citep{cobbe2021gsm8k,thoppilan2022lamda}, 
have also been applied to select suitable reasoning paths, they often rely on heuristic or trained smaller models. 
Recently,
\citet{li2022advance} sample different demonstrations and use step-by-step verification to filter out incorrect answers. 
However, obtaining step-level labels can be challenging, 
and using smaller models for judgment struggles to handle complex reasoning processes.
In contrast, our method fully utilizes the communication and decision-making capabilities of LLMs to reach the final answer, without the need for additional training and annotated data.

\subsection{Reasoning Path Refinement}
Although CoT~\citep{wei2022chain} effectively enhances the performance of LLMs in complex reasoning tasks, 
they remain susceptible to errors during the reasoning process,
leading to incorrect answers~\citep{bai2022constitutional, lyu2023faithful}.
To mitigate this issue, 
starting from the model’s own thoughts,
\citet{shinn2023reflexion} and \citet{madaan2023self} employ the model's own feedbacks and past mistakes to refine the reasoning process.
\citet{yao2023react} explore the synergies between reasoning chains and action plans.
For numerical problems,
\citet{zheng2023progressive} gradually guide models to the correct answer by using previously generated answers as hints.
With the aid of external knowledge, 
\citet{wang2023boosting} introduce chain-of-knowledge prompting that employs evidence triples to curb the generation of unfactual and unfaithful answers.
Taking model interactions into account, multi-agent debates~\citep{du2023improving,liang2023encouraging} have been introduced to enhance the factual accuracy of generated content and reduce fallacies and hallucinations.
EoT differs from these efforts as we prioritize enhancing the current reasoning process generated by a single model by incorporating the reasoning processes from other models as external insights through cross-model communication.

\begin{figure*}[t]
    \centering
 \includegraphics[width=0.8\linewidth]{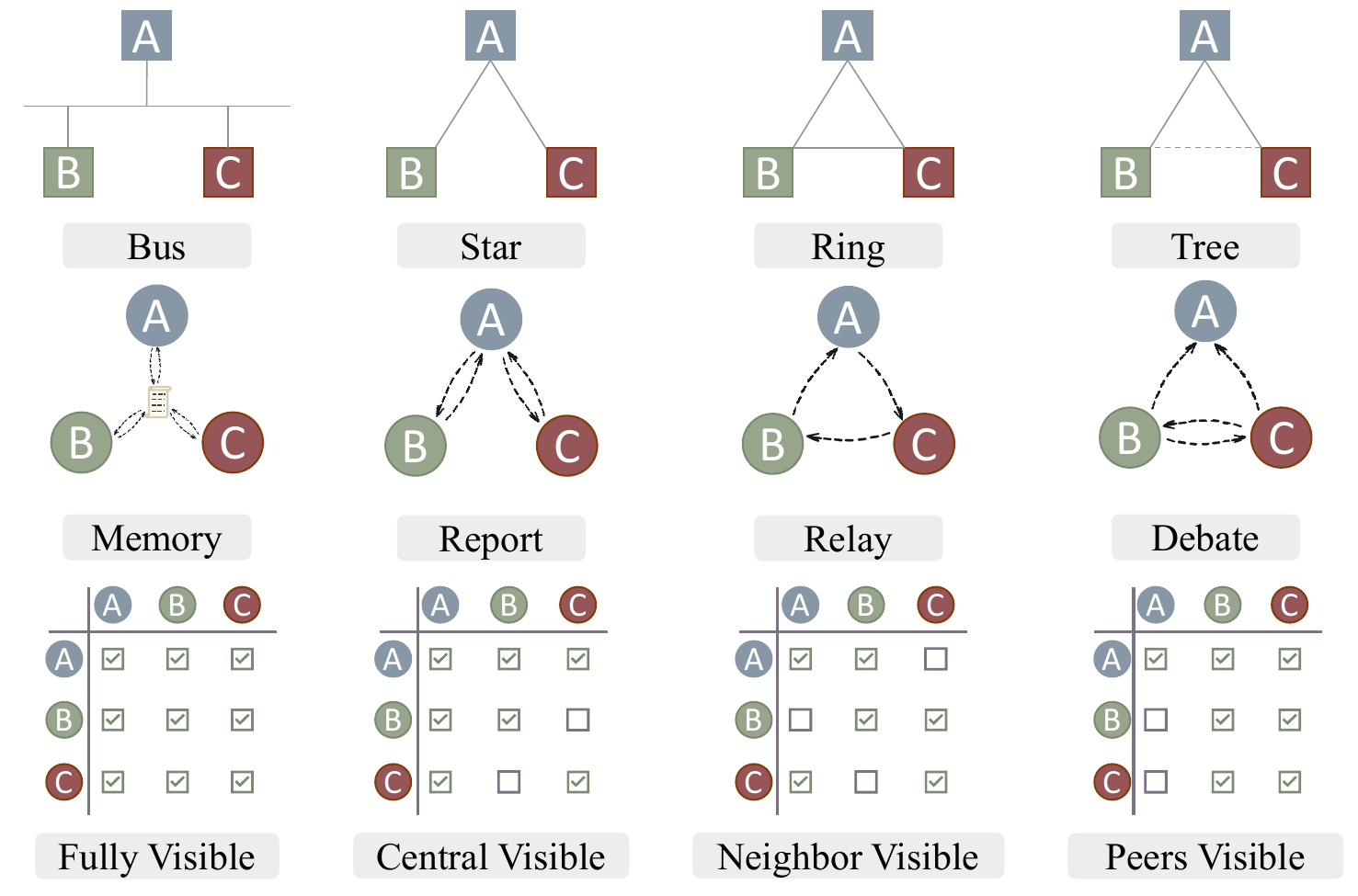}
    \vspace{-0.5em}
    \caption{Correspondence between communication paradigms and network topologies. The top row depicts four network topologies. The second row correlates these with the corresponding communication paradigms. The bottom row offers an  analysis of the communication volume associated with each paradigm. The horizontal axis represents the information that the node can receive, while the vertical axis indicates the information that the node can send.}
    \label{fig:framework}
    \vspace{-1em}
\end{figure*}

\section{Preliminary}
\label{sec:preliminary}
Firstly, we define the current methods that use LLMs to solve problems. We denote a LLM with a parameter size of \(\theta\) as \(p_{\theta}\), and the sequence length as \(t\), which includes tokens \([s_1,s_2,\dots,s_t]\). The LLM predicts the next token based on the prior tokens in the sequence. The probability of the \(s_i\) token is \(p_{\theta}(s_i|s_1,s_2,\dots,s_{i-1})\). Therefore, the probability of the whole sentence is \(p_{\theta}(s)=\prod_{i=1}^{t}p_{\theta}(s_i|s_{\leq i-1})\).

\vspace{-.2em}
\paragraph{Standard prompting.} Standard prompting involves deriving an answer \(a\) from a question \(q\) using \(p_{\theta}(a|q)\). In-Context Learning~\citep{brown2020language}
aims to improve LLMs performance by adding demonstrations \(\mathit{D}=\{d_1,d_2,\dots,d_n\}\) to the input, which can be expressed as \(p_{\theta}(a|\mathit{D},q)\).

\vspace{-.2em}
\paragraph{CoT prompting.} As identified by \citet{wei2022chain}, the incorporation of intermediate reasoning steps can improve the proficiency of LLMs in tackling complex reasoning challenges. To facilitate this, a rationale \(r_i\) is added to demonstration \(d_i=\{q_i,r_i,a_i\}\) to guide the LLMs in explicitly generating reasoning steps. \citet{fu2023complexcot} observe that using rationale \(r_i\) with more complex reasoning steps for demonstrations can further enhance the model's reasoning performance.

\vspace{-.2em}
\paragraph{Self-Consistency.} Self-Consistency method, introduced by \citet{wang2023sc}, effectively consolidates answers from multiple independent reasoning chains. This technique prioritizes the most commonly occurring answer, defined as \( a = \mathrm{argmax}_{a_i} f(a_i) \), where \( f(a_i) \) denotes the frequency of each answer \( a_i \). This approach enables the model to explore a broader range of reasoning pathways, thereby enhancing its reasoning ability. However, it remains constrained by the intrinsic limitations of LLMs' capabilities.

\vspace{-.2em}
\paragraph{Progressive-Hint Prompting.} Introduced by \citet{zheng2023progressive}, Progressive-Hint Prompting (PHP) leverages a sequence of historical answers \( \{a^{(1)}, a^{(2)}, \dots, a^{(j-1)}\}\) to enhance the current reasoning process \(r^{(j)}\) and facilitate the derivation of the subsequent answer \(a^{(j)}\).

\section{Methodology}
\label{sec:methodology}

We introduce Exchange-of-Thought (EoT), a novel framework designed to facilitate cross-model communication, allowing for the exchange of reasoning processes to integrate external insights. This innovative approach leverages the communicative abilities of LLMs to promote the sharing of rationale \(r\) and answer \(a\) among participating models, fostering a collaborative environment for thought and analysis. The implementation of EoT encounters three key challenges:

\begin{enumerate}
\vspace{-.3em}
\item How to identify the appropriate counterparts for model communication?
\vspace{-.3em}
\item What are the conditions for ceasing communication between models?
\vspace{-.3em}
\item How to minimize the influence of incorrect reasoning during the communication process?
\end{enumerate}

\subsection{Communication Paradigm}
\label{sec:communication-paradigm}

Inspired by network topology~\citep{bisht2015analytical} and intelligent agent communication~\citep{parsons2003argument},
we propose four communication paradigms to determine the counterparts for model communication.
As illustrated in Figure~\ref{fig:framework}, we propose Memory, Report, Relay, and Debate  communication paradigms each corresponding to the Bus, Star, Ring, and Tree network topologies, respectively. Assume in \(j\)-th round of communication, given a set of LLMs \(\{M\} = \{m_1, m_2, \dots, m_n\}\), the model \(m_i\) generates the corresponding rationale \(r_i^{(j)}\) and the answer \(a_i^{(j)}\) based on the \((r_K^{(j-1)}, a_K^{(j-1)})\), where \(K\) is the set from which model \(m_i\) can receive reasoning processes. In the first round, we use the CoT method proposed by~\citet{wei2022chain} to generate \((r^{(1)}, a^{(1)}) \sim P_{\theta}(r^{(1)}, a^{(1)}|\mathit{D},q)\).

\vspace{-.2em}
\paragraph{Memory.} Under the Memory paradigm, all models record their rationale \(r\) and answer \(a\) in a logbook, which is fully visible from all models. This means that in the \(j\)-th round, any model, such as model \(m_A\), can access the reasoning chains and answers from all models \({(r_{m}^{(j-1)}, a_{m}^{(j-1)})},\ m \in \{M\}\). As depicted in Figure~\ref{fig:framework}, this paradigm facilitates the fastest flow of information and also incurs the highest communication cost among all paradigms.

\vspace{-.2em}
\paragraph{Report.} Under the Report paradigm, we designate model \(m_A\) as the central node, which can obtain the rationale and answer from all other models \({(r_{m}^{(j-1)}, a_{m}^{(j-1)})},\ m \in \{M\} \backslash \{m_A\}\). Both \(m_B\) and \(m_C\) only receive information from \(m_A\) and do not interact with each other. Consequently, \(m_A\) plays a pivotal role in the communication process. This paradigm also allows for rapid information flow, but it demands a higher capacity for processing and analysis for the central node.

\vspace{-.2em}
\paragraph{Relay.} Under the Relay paradigm, we order the models by number and connect them in a circle. Each node is capable of receiving information from the preceding node and transmitting its own information to the subsequent node. For example, in the \(j\)-th round, \(m_A\) passes \((r_{A}^{(j-1)}, a_{A}^{(j-1)})\) to \(m_C\) and receives \((r_{B}^{(j-1)}, a_{B}^{(j-1)})\) from the previous round of \(m_B\). This distributed communication mode can reduce the demands on the information processing capacity of each node, but it may result in a slower flow of information.

\vspace{-.2em}
\paragraph{Debate.} We have adapted the tree topology to devise the Debate paradigm. This paradigm permits leaf nodes to exchange information with each other, while parent nodes are solely responsible for aggregating information, meaning that information flow is directed upward from child to parent. As illustrated in Figure~\ref{fig:framework}, \(m_B\) and \(m_C\), as child nodes, are able to communicate, whereas \(m_A\), as a parent node, can only receive information from its children. This communication paradigm strikes a balance between the model's information processing capacity and the speed of information flow. 



\subsection{Communication Volume}
\label{sec:communication-volume}
The last row of figure~\ref{fig:framework} displays the information that can be transmitted and received in different communication paradigms. The communication volume is measured by the number of messages received, assuming there are \( n \) models participating in the communication, with each node transmitting its information from the previous round to the next. 

In the Memory paradigm, every node receives information from all other nodes in the previous round, resulting in a communication volume of \( n^2 \). Any piece of information requires only one transmission to reach the corresponding node. 

Under the Report paradigm, the central node receives information from \( n-1 \) non-central nodes, while each of the \( n-1 \) non-central nodes receives information from the central node. In addition, each node can receive information from its previous round. Thus, the total communication volume is \( 3n - 2 \).  The transmission from a non-central node to another non-central node requires two transmissions, whereas sending to the central node requires only one. Thus, the average communication volume is calculated as \( 2 - \frac{2}{n} \).

Under the Relay paradigm, each node receives information from the preceding node and its own information from the last round, resulting in a communication volume of \( 2n \). Node \( i \) sends information to node \( i+1 \) in just one transmission, but sending to node \( i-1 \) requires \( n-1 \) transmissions. Therefore, the average propagation speed is \( \frac{n}{2} \).

In the Debate paradigm, nodes are assumed to form a full binary tree with a height of \( h = \lceil \log_{2}(n+1) \rceil \). The communication volume for each pair of child nodes is 4, and it is 3 for the parent node. Consequently, a subtree comprising two children and one parent has a communication volume of 7. The number of non-leaf nodes in a full binary tree is \( \frac{n-1}{2} \), leading to a total communication volume of \( \frac{7(n-1)}{2} \). Information under the same parent node requires only one transmission, whereas the information from the farthest nodes needs \( h-1 \) transmissions to converge at the root node. Thus, the communication speed \( \mathcal{S} = \frac{\Sigma^{h-1}_{i=1}2^{i-1}i}{2^{h-1}-1} \).

\begin{figure*}[t]
    \centering
 \includegraphics[width=1\linewidth]{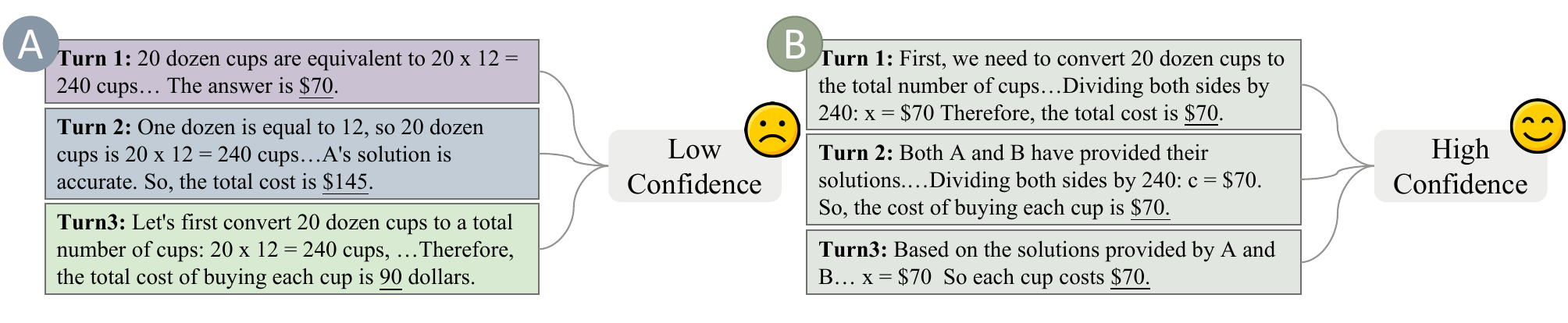}
    \vspace{-1em}
    \caption{An illustrative comparison between a confident model and an unconfident model. Model A generates three different answers over three communication rounds, indicating uncertainty about the answer, while Model B consistently adheres to a single answer.}
    \label{fig:EoT-confidence}
    \vspace{-1em}
\end{figure*}

\subsection{Termination Condition}
\label{sec:termination-condition}
Utilizing the models' current round outputs and the answers from previous rounds, we have devised two criteria for terminating communication: consistent output and majority consensus.

\vspace{-.2em}
\paragraph{Consistent Output Termination.} Inspired by \citet{zheng2023progressive}, we implement a consistent output termination in EoT. The termination condition is triggered when the output of model \(m_i\) in the \(j\)-th round is the same as the output in the \(j-1\)-th round, \(a_i^{(j)}=a_i^{(j-1)}\). In this case, \(m_i\) will stop receiving or sending information and exit the current communication.

\vspace{-.2em}
\paragraph{Majority Consensus Termination.} \citet{du2023improving} observed that LLMs can converge on a consensus after several rounds of debate, suggesting that LLMs fine-tuned with reinforcement learning from human feedback (RLHF)~\citep{ouyang2022training} are more likely to reach an agreement. Inspired by this finding, we propose the termination condition of majority rule, where LLMs cease communication with each other once a majority of them reach an agreement. This approach serves as a global termination condition, distinguishing it from the consistent output termination, which acts as a cessation criterion on an individual model basis.

\subsection{Confidence Evaluation}
\label{sec:confidence-evaluation}

An intriguing aspect of human behavior is that individuals are less likely to make mistakes when they are confident in their answers. Conversely, when uncertain about their answers, they become more susceptible to the influence of others' opinions. Additionally, as found by \citet{wang2023sc}, the likelihood of an answer being correct decreases as the generated results become more contradictory. Therefore, if a model's answers frequently change during communication, there is a high probability that these answers are incorrect.

We propose calculating the model's confidence based on the variation in responses.
This aids the recipient of the information in verifying its reliability, 
thereby safeguarding the problem-solving process from the disruption of erroneous information. 
Figure~\ref{fig:EoT-confidence} presents an illustrative example of a confident model and a non-confident model.

In a communication with $k$ rounds, model $m_i$ generates a set of answers \(\{a_i^{(1)}, \dots, a_i^{(k)}\}\). Let \(f(a_i) = \mathrm{max} ~\# \{a~|~a = a_i^{(j)} \}\) denote the number of the most frequently occurring answer from model \(m_i\). Consequently, we obtain the model's confidence level \(\ \mathcal{C}_i = \frac{f(a_{i})}{k}\) in the current round.


\section{Experiments}
\label{sec:experiment}
\definecolor{MorandiBlue}{RGB}{118,134,146}

\begin{table*}[t]
\centering
\label{tab:main_results}
\resizebox{\linewidth}{!}{
\begin{tabular}{lccccccc}
\toprule
{\textbf{Method}} & \textbf{GSM8K} & \textbf{MultiArith} & \textbf{SingleEQ} & \textbf{AddSub} & \textbf{AQuA} & \textbf{SVAMP} & {\textbf{Avg.}} \\
\midrule
\multicolumn{8}{c}{\textit{Single Reasoning Chain}} \\ 
\midrule
CoT  & 79.12\vari{0.50} & 97.27\vari{0.65} & 92.80\vari{0.27} & 86.23\vari{0.52} & 55.12\vari{1.03} & 79.52\vari{0.81} & 81.67 \\
ComplexCoT  & 79.32\vari{0.65} & {95.40}\vari{0.50} & {91.34}\vari{0.33} & 84.46\vari{0.86} & 56.46\vari{0.59} & 77.70\vari{0.54} & 80.78 \\
CoT (GPT-4) & \textbf{94.90} & 97.80 & 93.10 & \textbf{89.30} & \textbf{77.50} & \textbf{90.50} & 90.51 \\
\midrule
\multicolumn{8}{c}{\textit{Ensemble Methods}} \\ 
\midrule
CoT-SC(3) & 82.82\vari{0.32}  & 98.20\vari{0.43}  & 93.31\vari{0.12}  & 87.19\vari{0.47}  & 62.13\vari{1.30}  & 81.98\vari{0.49}  & 84.27 \\
CoT-SC(5) & 85.47\vari{0.52}  & 98.60\vari{0.08}  & 93.70\vari{0.25}  & 87.49\vari{0.38}  & 64.02\vari{0.95}  & 83.76\vari{0.81}  & 85.50 \\
CoT-SC(10) & 87.57\vari{0.27}  & 98.97\vari{0.12}  & 94.06\vari{0.36}  & 87.59\vari{0.58}  & 66.38\vari{1.72}  & 84.96\vari{0.33}  & 86.59 \\
ComplexCoT-SC(3) & 84.17\vari{0.67}  & 97.43\vari{0.31}  & 92.95\vari{0.53}  & 86.13\vari{0.74}  & 60.47\vari{1.55}  & 81.44\vari{0.79}  & 83.77 \\
ComplexCoT-SC(5) & 87.26\vari{0.33}  & 98.13\vari{0.22}  & 94.02\vari{0.29}  & 86.48\vari{0.61}  & 62.05\vari{2.40}  & 83.86\vari{0.92}  & 85.30 \\
ComplexCoT-SC(10) & 89.23\vari{0.31}  & 98.23\vari{0.37}  & 94.21\vari{0.16}  & 86.58\vari{0.58}  & 64.96\vari{1.91}  & 85.58\vari{0.87}  & 86.46 \\
PHP & 85.10 & 98.00 & 92.90 & 85.30 & 60.60 & 83.10 & 84.16 \\
\midrule
\multicolumn{8}{c}{\textit{Exchange-of-Thought}} \\ 
\midrule
EoT-Memory & \cellcolor{MorandiBlue!70}\underline{88.98\vari{0.89}}  & \cellcolor{MorandiBlue!10}98.80\vari{0.16}  & \cellcolor{MorandiBlue!30}94.09\vari{0.48}  & \cellcolor{MorandiBlue!30}87.65\vari{0.49}  & \cellcolor{MorandiBlue!10}69.37\vari{2.77}  & \cellcolor{MorandiBlue!10}84.28\vari{0.48}  & 87.20 \\
EoT-Report & \cellcolor{MorandiBlue!50}88.61\vari{0.83}  & \cellcolor{MorandiBlue!70}\underline{\textbf{99.03\vari{0.22}}}  & \cellcolor{MorandiBlue!10}94.06\vari{0.47}  & \cellcolor{MorandiBlue!70}\underline{87.95\vari{0.34}}  & \cellcolor{MorandiBlue!50}70.31\vari{2.19}  & \cellcolor{MorandiBlue!30}84.78\vari{0.75}  & 87.46 \\
EoT-Relay  & \cellcolor{MorandiBlue!10}88.42\vari{0.72}  & \cellcolor{MorandiBlue!50}98.97\vari{0.16}  & \cellcolor{MorandiBlue!50}94.13\vari{0.49}  & \cellcolor{MorandiBlue!10}87.59\vari{0.58}  & \cellcolor{MorandiBlue!70}\underline{70.87\vari{1.98}}  & \cellcolor{MorandiBlue!50}85.04\vari{0.31}  & 87.50 \\
EoT-Debate & \cellcolor{MorandiBlue!30}88.52\vari{0.76}  & \cellcolor{MorandiBlue!30}98.90\vari{0.17}  & \cellcolor{MorandiBlue!70}\underline{\textbf{94.25\vari{0.19}}}  & \cellcolor{MorandiBlue!50}87.70\vari{0.34}  & \cellcolor{MorandiBlue!30}69.69\vari{1.24}  & \cellcolor{MorandiBlue!70}\underline{85.10\vari{0.24}}  & 87.36 \\
\bottomrule
\end{tabular}
}
\vspace{-.5em}
\caption{Comparison of EoT performance with a series of strong baselines on mathematical reasoning tasks. The best results are highlighted in bold, while the best results among different EoT paradigms are underlined. The performance of different EoT communication paradigms is represented by varying colors, with darker shades indicating higher performance. The results for CoT (GPT-4) and PHP are reported from \citet{zheng2023progressive}.
}
\label{tab:mathematical-results}
\vspace{-1em}
\end{table*}
\begin{figure*}[htb]
    \centering
    \begin{subfigure}[t]{0.24\textwidth}
        \includegraphics[width=\linewidth]{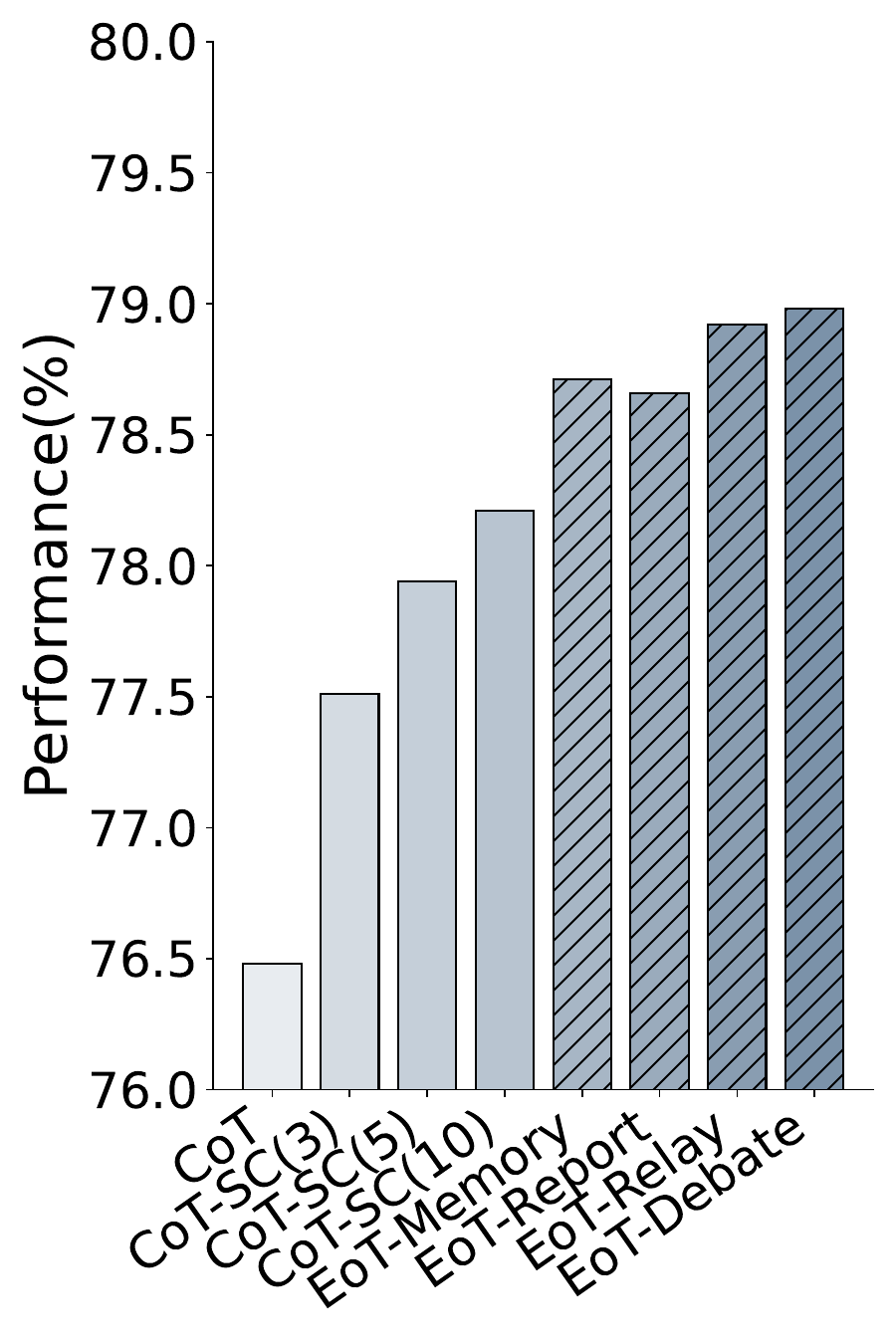}
        \vspace{-1.7em}
        \caption{CSQA.}
        \label{fig:csqa-result}
    \end{subfigure}
    \hfill
    \begin{subfigure}[t]{0.24\textwidth}
        \includegraphics[width=\linewidth]{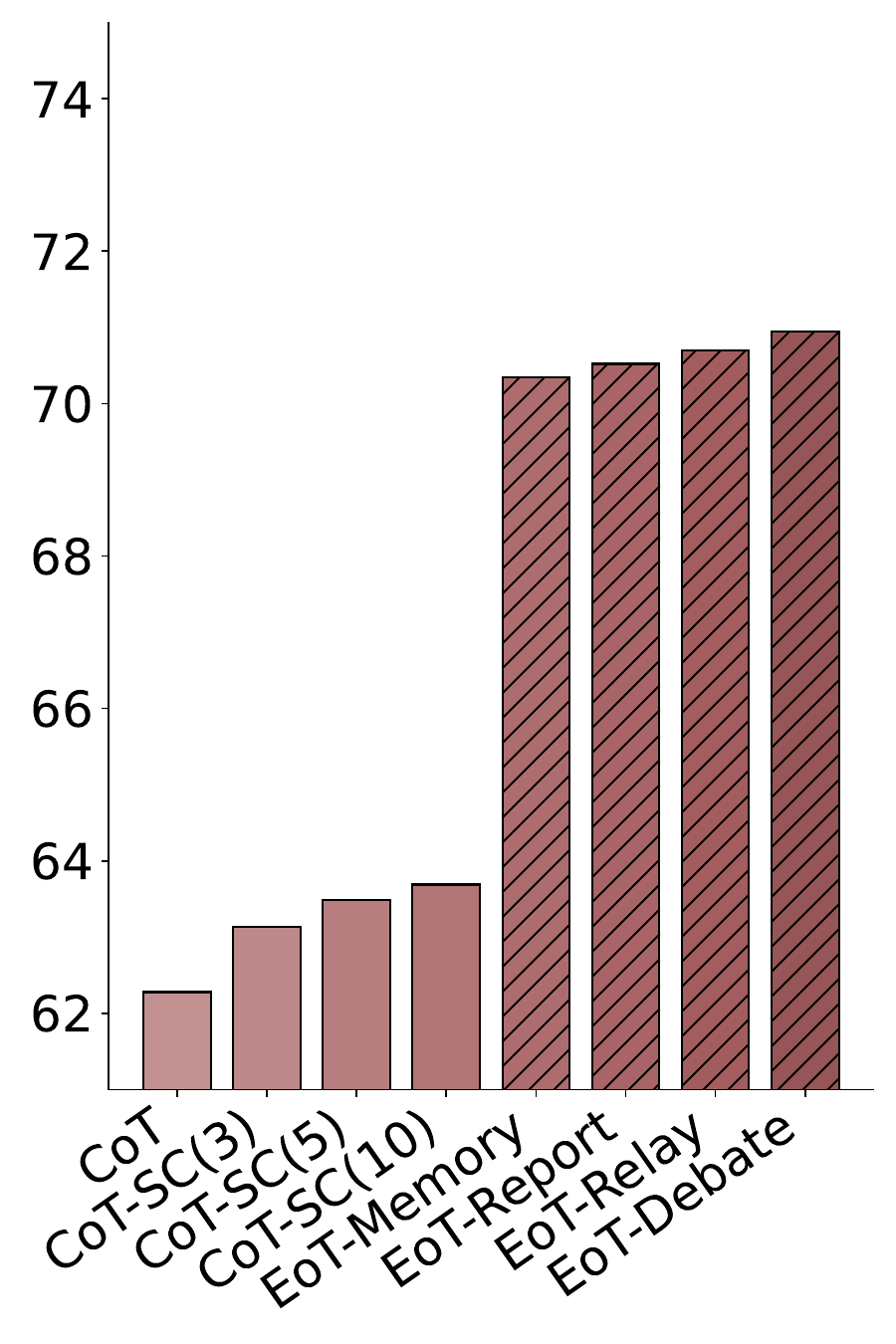}
        \vspace{-1.7em}
        \caption{StrategyQA.}
        \label{fig:strategyqa-result}
    \end{subfigure}
    \hfill
    \begin{subfigure}[t]{0.24\textwidth}
        \includegraphics[width=\linewidth]{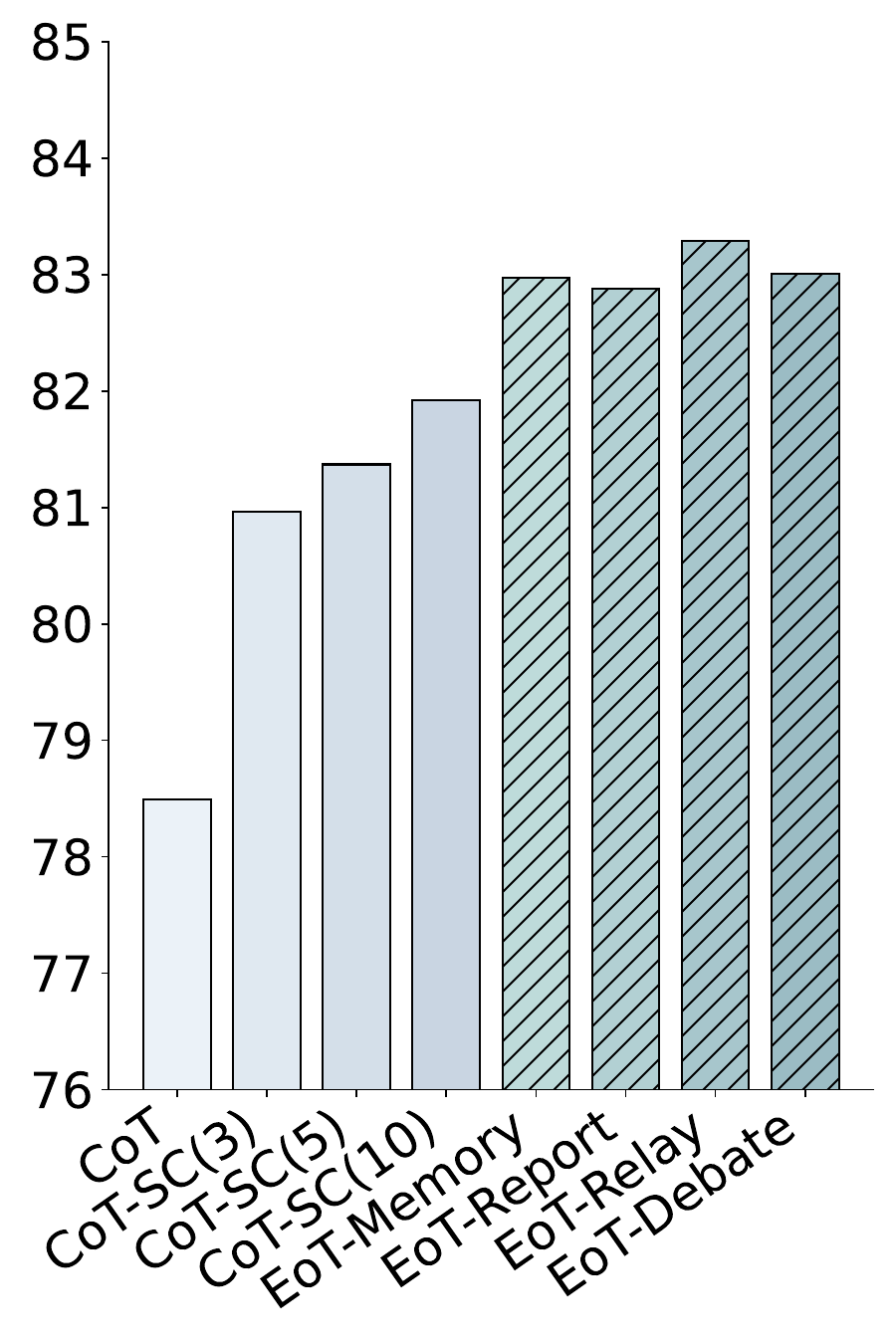}
        \vspace{-1.7em}
        \caption{Peguins.}
        \label{fig:penguins-result}
    \end{subfigure}
    \hfill
    \begin{subfigure}[t]{0.24\textwidth}
        \includegraphics[width=\linewidth]{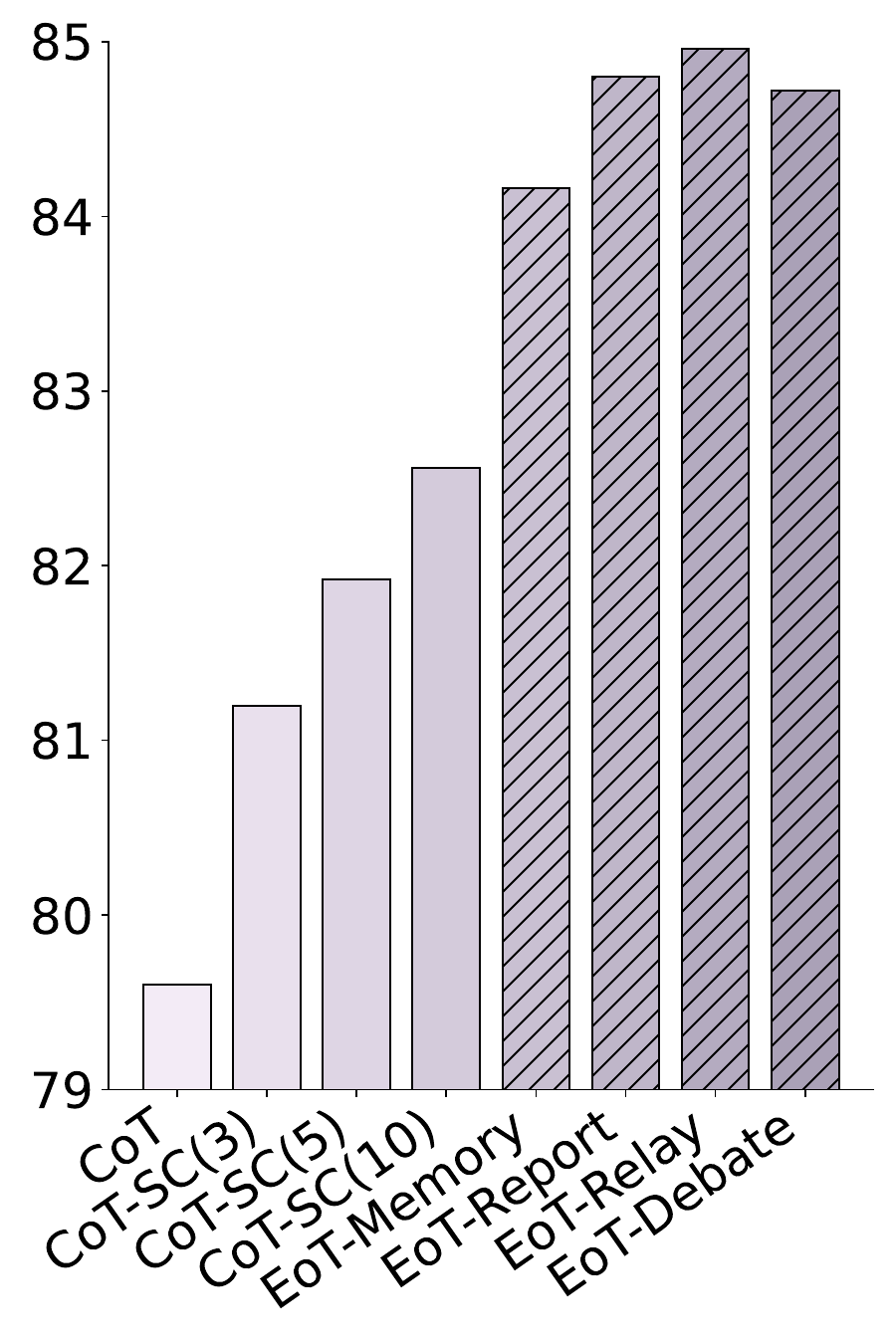}
        \vspace{-1.7em}
        \caption{Date Understanding.}
        \label{fig:date-result}
    \end{subfigure}
    \vspace{-.5em}
    \caption{Comparison of EoT with CoT and CoT-SC methods in commonsense and symbolic reasoning tasks.}
    \vspace{-.5em}
    \label{fig:comparison}
\end{figure*}

\subsection{Experimental Setups}
\label{sec:exp-experiment-setups}
\paragraph{Tasks and Datasets.} In our experiments, we evaluated the performance of EoT across three complex reasoning tasks:
\textbf{(1) Mathematical Reasoning}: This involves six datasets, which includes GSM8K~\citep{cobbe2021gsm8k}, MultiArith~\citep{roy2015multiarith}, SingleEQ~\citep{koncel2015parsing}, AddSub~\citep{hosseini2014addsub}, AQuA~\citep{ling2017aqua}, and SVAMP~\citep{patel2021svamp}.
\textbf{(2) Commonsense Reasoning}: We utilize the CommonsenseQA(CSQA; \citealp{talmor2019commonsenseqa}) and StrategyQA~\citep{geva2021strategyqa}.
\textbf{(3) Symbolic Reasoning}: We employ two datasets from BigBench~\citep{srivastava2023bb, suzgun2023bbh}, namely Penguins in a Table (Penguins) and Date Understanding. In Appendix~\ref{appendix:app-datasets}, we provide a detailed description and statistics of the datasets.

\vspace{-.2em}
\paragraph{Baselines.} 
\label{sec:exp-baselines}
We compare EoT with a series of strong baselines, which include 
(1) Chain-of-Thought prompting (CoT; \citealp{wei2022chain}), 
(2) Complexity-based prompting (ComplexCoT; \citealp{fu2023complexcot}), 
(3) Self-Consistency (SC; \citealp{wang2023sc}),  
(4) Progressive Hint Prompting (PHP; \citealp{zheng2023progressive}). 
Specifically, CoT and ComplexCoT are prompting methods, 
while SC and PHP are reasoning chain ensemble methods. 
For simplicity in notation, we use ``CoT-SC(10)'' to denote the approach that employs the CoT prompt method to sample 10 reasoning chains and then utilize the SC method to select the answer.

\vspace{-.2em}
\label{sec:exp-implemention-details}
\paragraph{Implementation Details.} We access the GPT models through the OpenAI API. In the main experiments, we employ \turbo (\turbon) and \gpt (\gptn) to evaluate the effectiveness of EoT in comparison to other strong baselines. We set the temperature at 1 during the generation. The prompts for CoT and ComplexCoT are sourced from \citet{wei2022chain} and \citet{fu2023complexcot}. By default, we employ three \turbo to engage in the EoT communication. We apply the majority consensus termination and confidence evaluation, selecting the majority answer as the final outcome. Taking into account the impact of temperature, we report the average performance and standard deviation across five runs. Additionally, in Section~\ref{sec:exp-discussions}, to further validate the performance of different LLMs on EoT, we incorporate the \claude model. 
The further implementation details are listed in Appendix~\ref{appendix:app-implementation_details}.

\begin{figure*}[ht]
    \centering
    \begin{minipage}[t]{0.45\textwidth}
        \includegraphics[width=\linewidth]{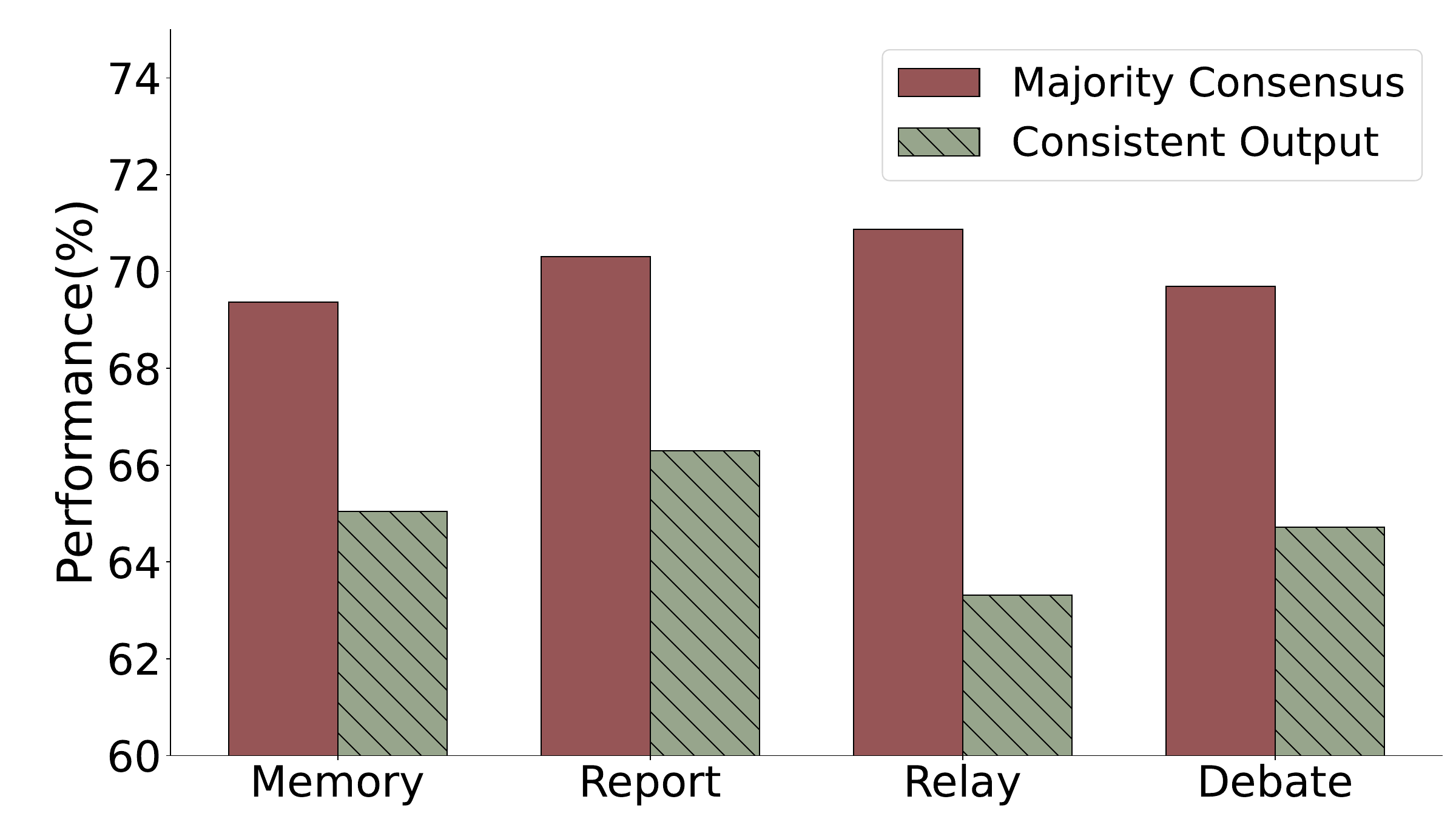}
        \caption{Comparison of consistent output termination and majority consensus termination on AQuA.}
        \label{fig:termination-condition}
    \end{minipage}
    \hfill
    \begin{minipage}[t]{0.45\textwidth}
        \includegraphics[width=\linewidth]{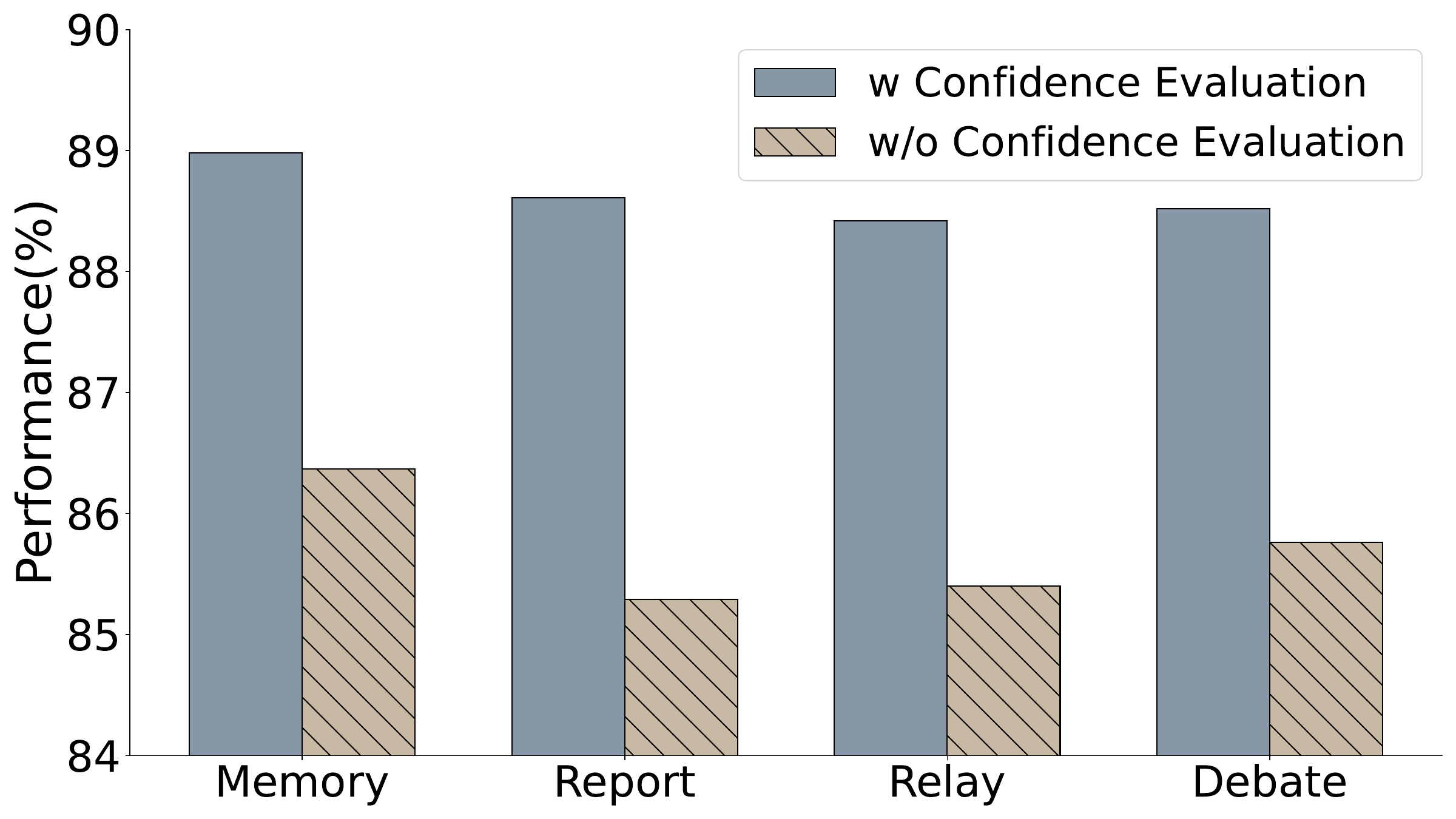}
        \caption{The impact of employing confidence evaluation on accuracy in the GSM8K dataset.}
        \label{fig:confidence-evaluation}
    \end{minipage}
    \vspace{-1em}
\end{figure*}

\begin{figure*}[t]
    \centering
    \begin{minipage}[t]{0.45\textwidth}
        \includegraphics[width=\linewidth]{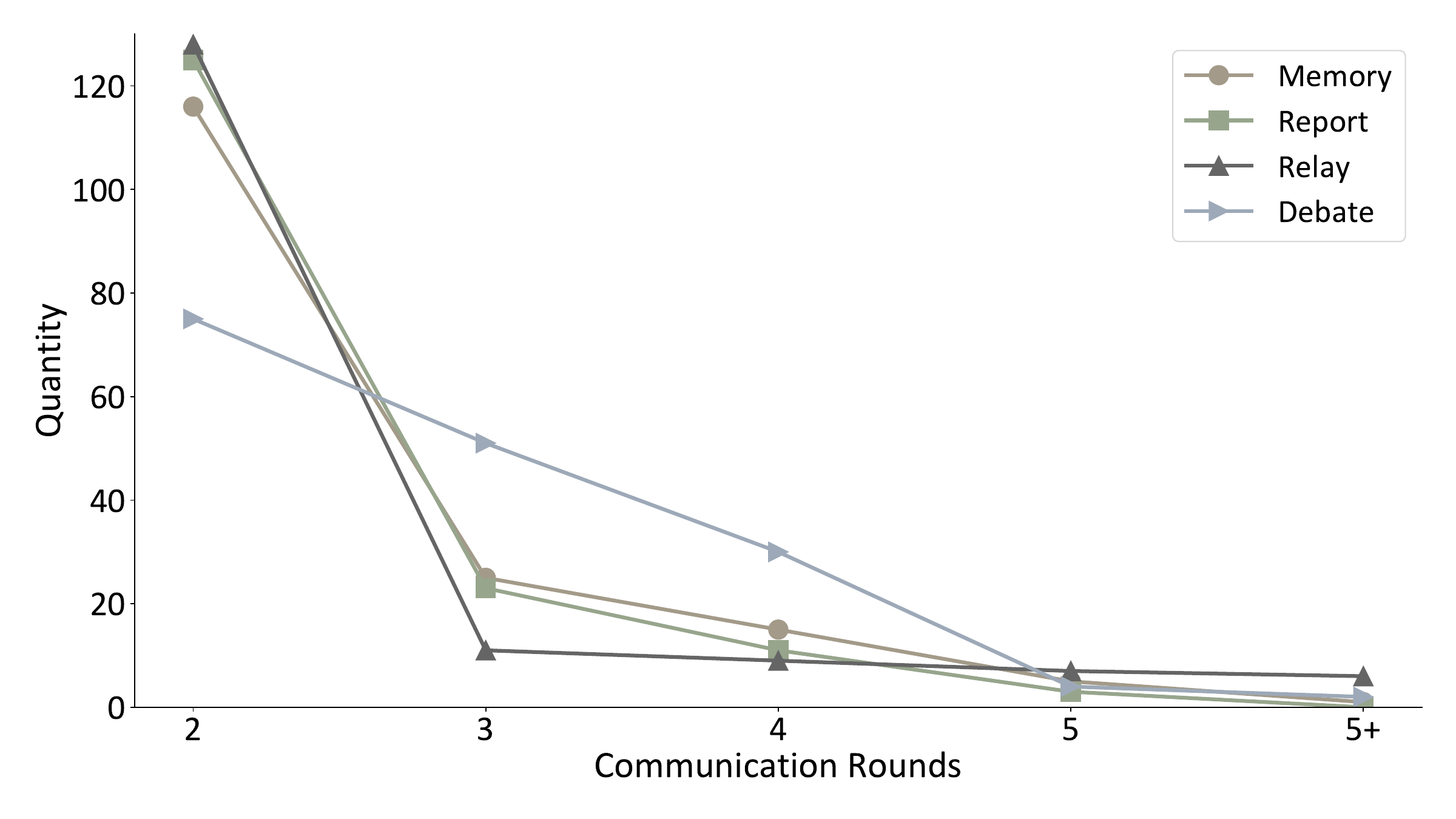}
        \caption{Number of communication rounds required to reach termination condition on SVAMP.}
        \label{fig:round-analysis}
    \end{minipage}
    \hfill
    \begin{minipage}[t]{0.5\textwidth}
        \includegraphics[width=\linewidth]{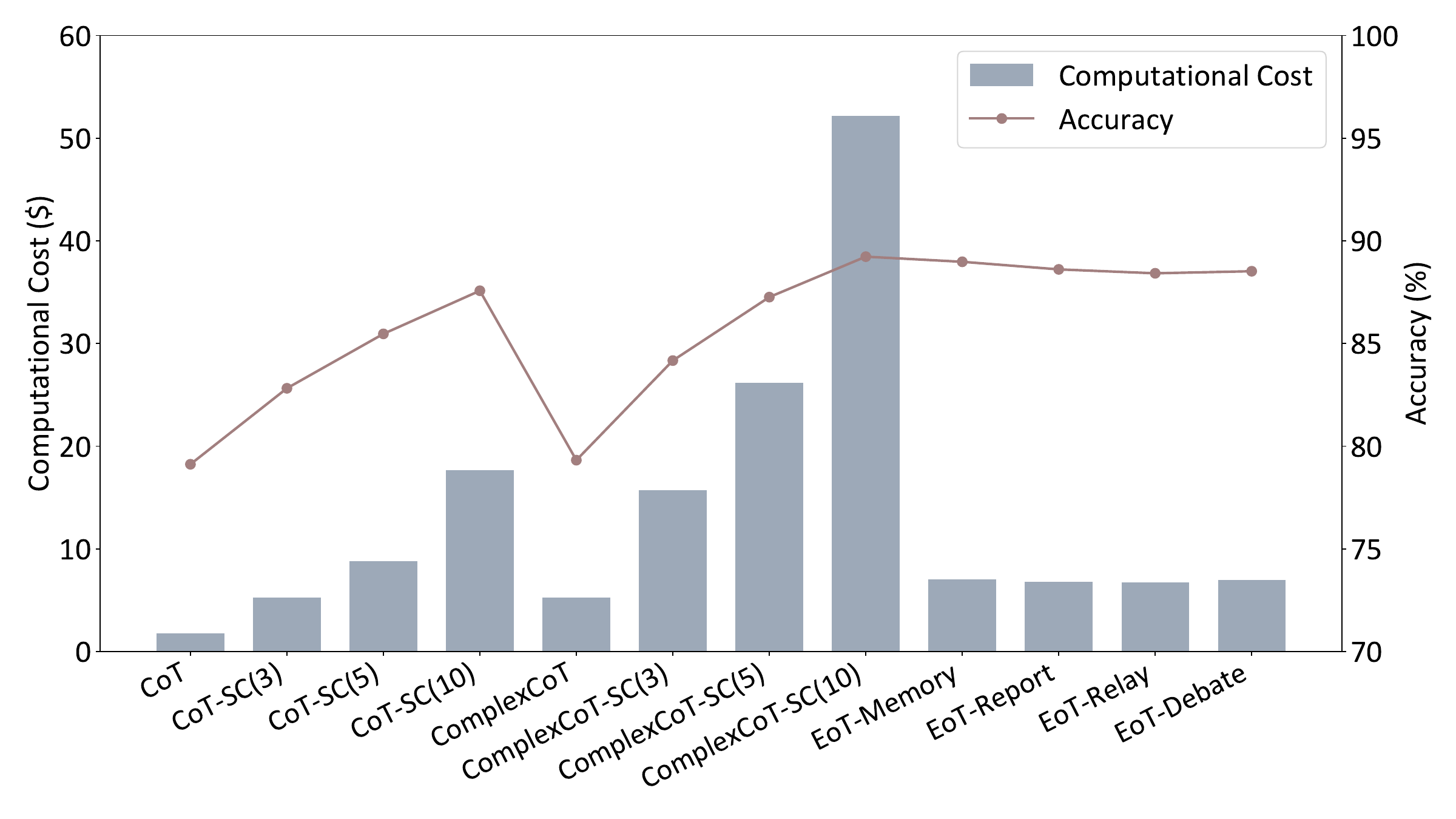}
        \caption{Performance and associated costs of different methods in the GSM8K dataset.}
        \label{fig:cost-analysis}
    \end{minipage}
    \vspace{-1em}
\end{figure*}

\begin{figure*}[t]
    \centering
    \begin{minipage}[t]{0.485\textwidth}
        \includegraphics[width=\linewidth]{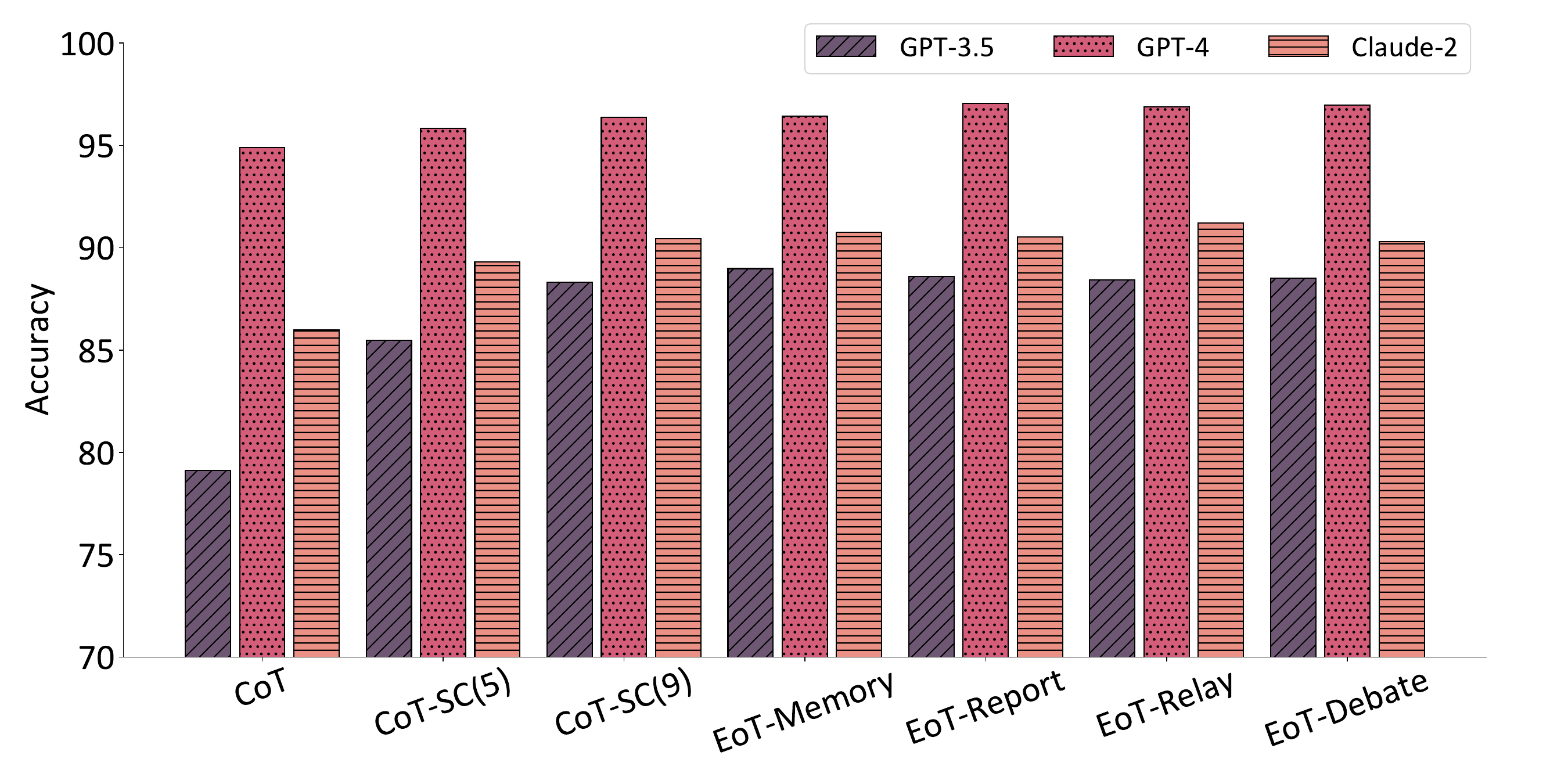}
        \caption{Comparison of EoT with CoT and CoT-SC methods using different LLMs as backbones on GSM8K.}
        \label{fig:model-analysis}
    \end{minipage}
    \hfill
    \begin{minipage}[t]{0.485\textwidth}
        \includegraphics[width=\linewidth]{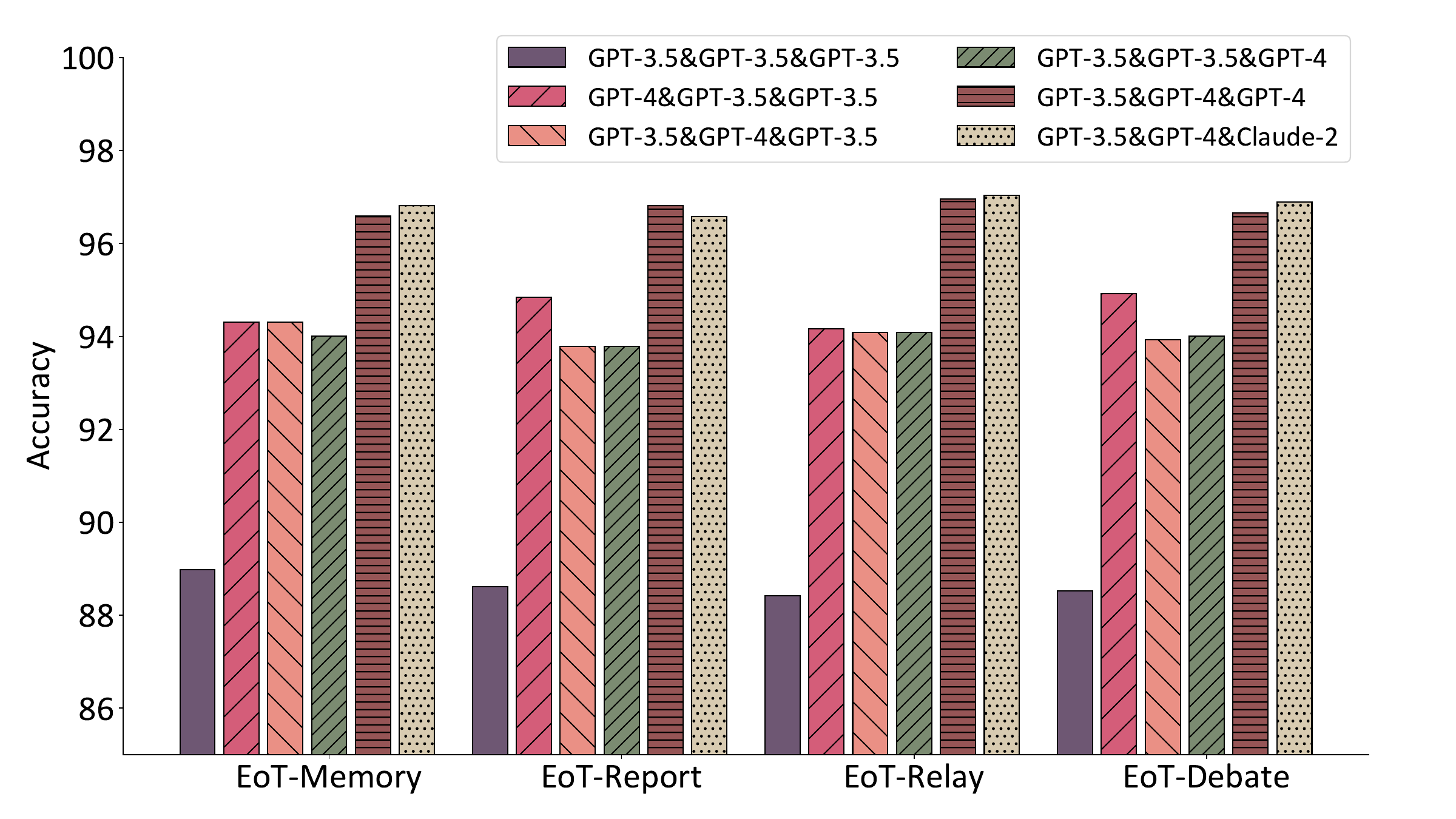}
        \caption{Effect of different node positions for LLMs on accuracy in the GSM8K Dataset.}
        \label{fig:position-analysis}
    \end{minipage}
    \vspace{-1em}
\end{figure*}

\subsection{Performance of EoT}
\label{sec:exp-performance-of-EoT}

\paragraph{Mathematical Reasoning.} According to the results presented in Table ~\ref{tab:mathematical-results}, the four communication paradigms of EoT have shown significant improvement over both CoT and ComplexCoT in mathematical reasoning tasks.
Compared to the currently strongest baseline method, PHP, the Memory, Report, Relay, and Debate paradigms have respectively increased the average performance by 3.04\%, 3.30\%, 3.34\%, and 3.20\%. 
EoT comprehensively outperforms CoT-SC(5), achieving performance comparable to, or even surpassing, that of CoT-SC(10).
When compared to the current best LLM \gptn, three \turbon with EoT surpassed a single \gptn with CoT on the MultiArith and SingleEQ datasets. 
This indicates that through cross-model communication and collaboration, three less capable models can compensate for their individual weaknesses and outperform more powerful model, showcasing the potential of EoT to enhance model capabilities and address inherent shortcomings by incorporating external insights.

\vspace{-.2em}
\paragraph{Commonsense Reasoning.} 
The comparison of EoT with CoT and CoT-SC methods on commonsense reasoning tasks is illustrated in Figures~\ref{fig:csqa-result} and~\ref{fig:strategyqa-result}. EoT significantly outperforms CoT. Specifically, on the StrategyQA dataset, Memory, Report, Relay, and Debate respectively achieve improvements of 8.06\%, 8.24\%, 8.42\%, and 8.67\% compared to CoT. Similar significant gains are observed on the CSQA dataset. Furthermore, across both commonsense reasoning tasks, all four paradigms outperform the CoT-SC(10) method, which samples 10 reasoning chains, demonstrating the superior performance of EoT.

\vspace{-.2em}
\paragraph{Symbolic Reasoning.} 
Figures~\ref{fig:penguins-result} and \ref{fig:date-result} compare the performance of EoT with CoT and CoT-SC methods on symbolic reasoning tasks. On the Penguins dataset, the Memory, Report, Relay, and Debate paradigms of EoT achieve improvements of 2.01\%, 1.92\%, 2.33\%, and 2.05\% respectively, compared to the CoT-SC(3) method which samples 3 reasoning chains. On the Date Understanding dataset, the performance gains of EoT are even more pronounced, with all four paradigms showing an average improvement of 2.1\% over CoT-SC(10).

\subsection{Discussions}
\label{sec:exp-discussions}

\paragraph{Communication Paradigm.} We propose four communication paradigms and analyze the communication volumes in Section~\ref{sec:communication-paradigm} and Section~\ref{sec:communication-volume}. In the results illustrated in Table~\ref{tab:mathematical-results}, we observe that different communication paradigms have their respective strengths. For instance, Report performs best on MultiArith and AddSub, while Debate achieves optimal performance on SingleEQ and SVAMP. This indicates that various communication paradigms are well-suited for different scenarios.

\vspace{-.2em}
\paragraph{Termination Condition.}
In Figure~\ref{fig:termination-condition}, we analyze the performance of the two termination conditions we propose in Section~\ref{sec:termination-condition} on the AQuA dataset. Compared to consistent output termination, majority consensus termination improved by 4.33\%, 4.01\%, 7.56\%, and 4.97\% under the Memory, Report, Relay, and Debate paradigms, respectively. Under consistent output termination, there is no mechanism for collective negotiation, and individual models are prone to premature exit due to degeneration~\citep{su2022a}. Therefore, majority consensus termination is more suitable for scenarios involving multiple model communication.

\vspace{-.2em}
\paragraph{Confidence Evaluation.}
We conduct ablation experiments on the GSM8K dataset for confidence evaluation. As shown in Figure~\ref{fig:confidence-evaluation}, across four communication paradigms, confidence evaluation show an average improvement of 2.92\% compared to the baseline. The introduction of confidence evaluation enables the model to consider the other model's confidence prior~\cite{zhang2023wisdom} during communication, facilitating its decision to accept the other model's reasoning chains at an earlier stage, thereby effectively mitigating the interference of incorrect reasoning chains.

\vspace{-.2em}
\paragraph{Round Analysis.}
As illustrated in Figure~\ref{fig:round-analysis}, we analyze the number of communication rounds to satisfy termination condition in the SVAMP dataset. For the majority of samples, consensus on the answer can be reached within three rounds of communication. \citet{wang2023sc} obverse that answer consistency is proportional to accuracy. EoT enables models to engage in a greater number of exchanges and discussions on questions where consensus is challenging to achieve. Consequently, a minority of difficult cases necessitate communication extending beyond five rounds.

\vspace{-.2em}
\paragraph{Cost Analysis.}
A potential concern is the computational expense incurred by EoT. In Figure~\ref{fig:cost-analysis}, we compare the performance and computational costs of CoT-SC, ComplexCoT-SC, and EoT methods. Compared to CoT-SC(5), EoT reduces costs by 20\% while enhancing performance by 3\%. EoT achieves comparable performance to ComplexCoT-SC(10) at only one-seventh of its cost. Since the majority of samples conclude communication within three rounds, EoT does not impose a significant computational burden. By facilitating the exchange of external insights between models, EoT effectively enhances model performance, demonstrating a cost-effective advantage.

\vspace{-.2em}
\paragraph{Model Applicability.}
In Figure~\ref{fig:model-analysis}, we analyze the performance of EoT when applied to different LLMs. EoT, compared to CoT-SC(5), shows performance improvements of 3.2\% on \turbon, 1.0\% on \gptn, and 1.4\% on \claude, indicating that EoT is adaptable to various LLMs and effectively boosts performance across multiple LLMs.

\vspace{-.2em}
\paragraph{Position Analysis.}
In Figure~\ref{fig:position-analysis}, we investigate the impact of different LLMs occupying different node positions on performance.
Notably, positioning the more powerful \gptn as the central node in the Report paradigm yields a performance increase of over 1\% compared to when \gptn serves as a non-central node. In the Debate paradigm, \gptn as a parent node outperforms \gptn as a child node by 0.9\%. The location of \gptn has a negligible effect on the decentralized Relay and Memory paradigms. Additionally, a configuration with two \gptn models and one \turbon significantly outperforms one with two \turbon models and one \gptn, underscoring that incorporating more superior models can further enhance EoT's performance. The combination of \turbon, \gptn, and \claude achieves performance close to or exceeding that of two \gptn with one \turbon, suggesting that model diversity can effectively boost EoT's effectiveness, aligning with the ensemble theory~\citep{kuncheva2003measures} that diversity among models can improve performance.

\section{Conclusion}
We introduce Exchange-of-Thought (EoT), a novel framework that enriches models with external insights through cross-model communication. We develop four communication paradigms and conduct a thorough analysis of the communication volume and information propagation speed. To safeguard against the disruption of incorrect reasoning processes, we design a confidence evaluation mechanism. Experiment on mathematical, commonsense, and symbolic reasoning tasks demonstrates that EoT surpasses a series of strong baselines while also offering a cost advantage. Further analysis reveals that EoT is adaptable to various models, and the participation of a more diverse range of models can further enhance the performance of EoT.

\section*{Ethics Statement}
The EoT method presented in this paper does not require the collection or utilization of any personal information. The prompts we have designed and employed are free from personal data and avoid language that discriminates against individuals or groups. We have conducted a comprehensive research of the licenses for the datasets used in this paper, as detailed in Appendix~\ref{appendix:app-datasets}, and have ensured that our work complies with all the licensing requirements of these datasets.

\section*{Acknowledgement}
This work was supported by the National Key Research and Development Program of China (No.2022CSJGG0801), National Natural Science Foundation of China (No.62022027). We extend our sincerest gratitude to the reviewers for their insightful comments and suggestions, which have been instrumental in enhancing the quality of this manuscript.

\bibliography{anthology,custom}
\bibliographystyle{acl_natbib}

\appendix

\section{Limitations and Broader Impacts}
\label{appendix:app-limitations}

\begin{table*}[t]
\centering
\footnotesize
\begin{tabular}{l|c|c|c|c|c}
\toprule
\textsc{Dataset} & \textsc{Reasoning Task} & \textsc{Ans Type} & \textsc{\# Prompt} & \textsc{\# Test} & \textsc{License} \\
\midrule
\href{https://github.com/openai/grade-school-math}{GSM8K}~\citep{cobbe2021gsm8k} & Mathematical & Number & 8 & 1,319 & MIT License \\
\href{https://github.com/wangxr14/Algebraic-Word-Problem-Solver}{MultiArith}~\citep{roy2015multiarith} & Mathematical & Number & 8 & 600 & Unspecified \\
\href{https://gitlab.cs.washington.edu/ALGES/TACL2015}{SingleEq}~\citep{koncel2016mawps} & Mathematical & Number & 8 & 508 & Unspecified \\
\href{https://github.com/wangxr14/Algebraic-Word-Problem-Solver}{AddSub}~\citep{hosseini2014addsub} & Mathematical & Number & 8 & 395 & Unspecified \\
\href{https://github.com/deepmind/AQuA}{AQUA}~\citep{ling2017aqua} & Mathematical & Multi-choice & 4 & 254 & Apache-2.0 \\
\href{https://github.com/arkilpatel/SVAMP}{SVAMP}~\citep{patel2021svamp} & Mathematical & Number & 8 & 1,000 & MIT License \\

\href{https://www.tau-nlp.sites.tau.ac.il/commonsenseqa}{CommonsenseQA}~\citep{talmor2019commonsenseqa}  & Commonsense & Multi-choice & 7 & 1,221 &  Unspecified \\
\href{https://github.com/eladsegal/strategyqa}{StrategyQA}~\citep{geva2021strategyqa} & Commonsense & T/F & 6 & 2,290 &  MIT license \\

\href{https://github.com/suzgunmirac/BIG-Bench-Hard}{Date Understanding}~\citep{suzgun2023bbh} & Symbolic & Multi-choice & 3 & 369 &  MIT license \\
\href{https://github.com/suzgunmirac/BIG-Bench-Hard}{Penguins in a Table}~\citep{suzgun2023bbh} & Symbolic & Multi-choice & 3 & 146 &  MIT license \\

\bottomrule
\end{tabular}
\caption{Detailed statistics of the datasets utilized in our experiment. \textsc{Ans Type} indicates the form of the answer. \textsc{\# Prompt} represent the count of chain-of-thought exemplars employed as few-shot prompts for each task. \textsc{\# Test} indicates the quantity of samples contained within each dataset.
}
\label{tab:dataset_statistic}
\vspace{-1em}
\end{table*}

Given the current constraints in communication and analytical capacities of open-source models~\citep{fu2023improving}, as well as their substantial computational resource requirements~\citep{touvron2023llama2,  chowdhery2022palm}, we have not included these models in our experiment at this stage. However, we posit that open-source models with advanced comprehension and communication skills have the potential to match or even exceed the performance of commercial models~\citep{openai2023gpt4, ouyang2022training, chowdhery2022palm}, through the collaborative exchange of insights.

A critical factor in model communication is the handling of long text. The current context windows of these models limit our ability to incorporate a broader range of models in the communication process. Recent works~\citep[][\emph{inter alia}]{liu2023scaling, xiao2023efficient, wang2023augmenting, tworkowski2023focused, chen2023extending, ratner-etal-2023-parallel} have begun to overcome this limitation by equipping models with the ability to process longer texts, laying the foundation for increasing the number of models involved in communication. In addition, our experiments indicate that model communication can achieve effective performance with reduced computational resources, aligning with the sustainable development goals of AI community~\citep{van2021sustainable, wu2022sustainable}.

Furthermore, the concept of AI learning from each other to foster collective improvement is a focal point of current research~\citep{bai2022constitutional, ponnusamy2022feedback, lee2023rlaif}. Our aim and aspiration is to cultivate a collective intelligence among large language models~\citep{ha2022collective}. This approach not only optimizes individual model performance but also contributes to the broader AI research community's pursuit of more advanced, collaborative AI systems.

\section{Datasets and Evaluation Metrics}
\label{appendix:app-datasets}
\vspace{-.2em}
\paragraph{Datasets} In Table~\ref{tab:dataset_statistic}, we meticulously detail the specifics and statistics of each dataset employed in our experiments. This includes the data source, task type, answer type, the number of prompt samples used, the total number of test samples, as well as the licenses pertaining to each dataset.

\vspace{-.2em}
\paragraph{Evaluation Metrics} Accuracy is used as the metric for evaluation in our study. For datasets where the answer is numerical, we employ regular expressions to extract the number following the phrase ``the answer is'' and perform a numerical comparison with the provided answer. For datasets with multiple-choice and true/false questions, accuracy is calculated by checking if the option extracted from the response matches the correct answer.

In the main experiment, all test samples are used for evaluation. In the analysis part, due to rate limits and cost considerations, we set an upper limit on the sample size. Consequently, a maximum of 1,000 samples are utilized for each run.

\section{Implementation Details}
\label{appendix:app-implementation_details}
\vspace{-.2em}
\paragraph{Confidence Evaluation.}  Considering that confidence evaluation requires historical answers for reference, we begin incorporating the confidence information into the prompts from the second round of communication. Specifically, after calculating \( C_i \) using the method described in Section~\ref{sec:confidence-evaluation}, we preface the solution with ``\( \mathcal{M}_i \)'s confidence in this solution is \( \mathcal{C}_i \)'', where \( \mathcal{M}_i \) is the character name.

\vspace{-.2em}
\paragraph{Termination Condition.}
For the consistent output termination condition, a minimum of two rounds of communication is necessary, as it requires the model's answer from the previous round. Given that only three models are involved in the EoT communication, the exit of a single model reduces the interaction to a dialogue between the remaining two, potentially impeding their communication. Therefore, if a single model exits, we terminate the communication and select the exiting model's answer as the final result.

In the case of majority consensus termination, if the answers from all three models align in the first round, we deem further communication unnecessary and end the exchange. Given that only three models are involved in the communication, an exit based on two models holding the same incorrect answer could lead to an inaccurate final result. Therefore, during the initial five rounds, we require a unanimous agreement among all models before ceasing communication. If a consensus is not reached after five rounds, the majority answer will be adopted as the final outcome.

\vspace{-.2em}
\paragraph{Computation Cost.} Computational costs are calculated based on OpenAI's official pricing for \turbo, which is computed as \(\text{Input Tokens} \times 0.0015 / 1000 + \text{Output Tokens} \times 0.002 / 1000\).

\section{EoT Prompts}
\label{appendix:prompts}
During the EoT communication process, we assign different roles to the models. Table~\ref{table:role-prompt} displays the prompts for each role, wherein we have models A, B, and C take on the personas of Kitty, Ben, and Peter, three high school students, to facilitate the communication. The specific prompts for different datasets can be found in our \href{https://github.com/yinzhangyue/EoT}{Github} repository.

\begin{table*}[ht]
\begin{tabular}{p{13cm}}
\toprule
\midrule
\underline{\textbf{\textsc{Character Prompts}}} \\

\begin{minipage}[b]{0.16\columnwidth}
    \centering
    \raisebox{-.1in}{\includegraphics[width=\linewidth]{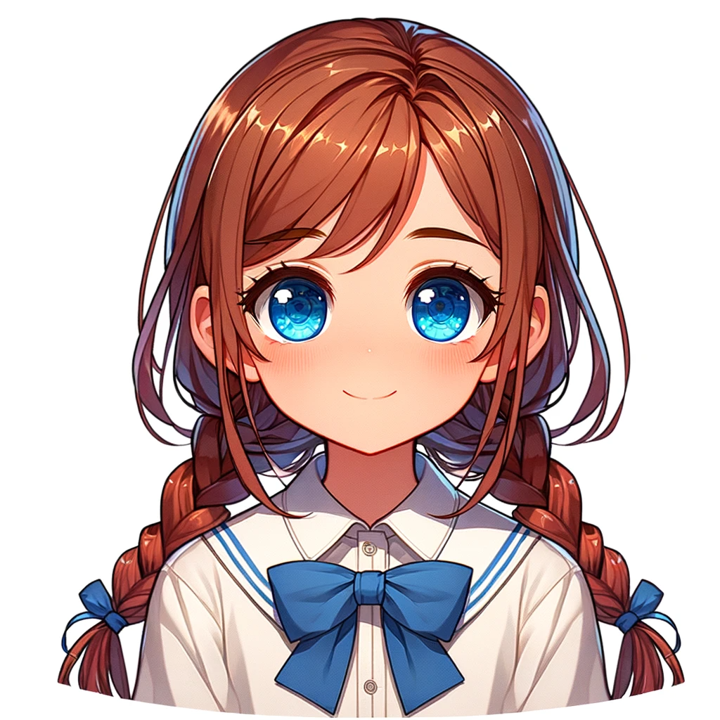}}
\end{minipage}
\textbf{Kitty}: ``You are Kitty, a high school student admired for your attentiveness and detail-oriented nature. Your friends often rely on you to catch details they might have missed in their work. Your task is to carefully analyze the presented math problem, apply your attentive skills, and piece together a detailed solution. Afterward, you'll have the opportunity to review the solutions provided by your friends, offering insights and suggestions. Your careful revisions will help all of you to enhance your understanding and arrive at the most accurate solutions possible.'' \\

\vspace{-1mm}
\begin{minipage}[b]{0.16\columnwidth}
    \centering
    \raisebox{-.1in}{\includegraphics[width=\linewidth]{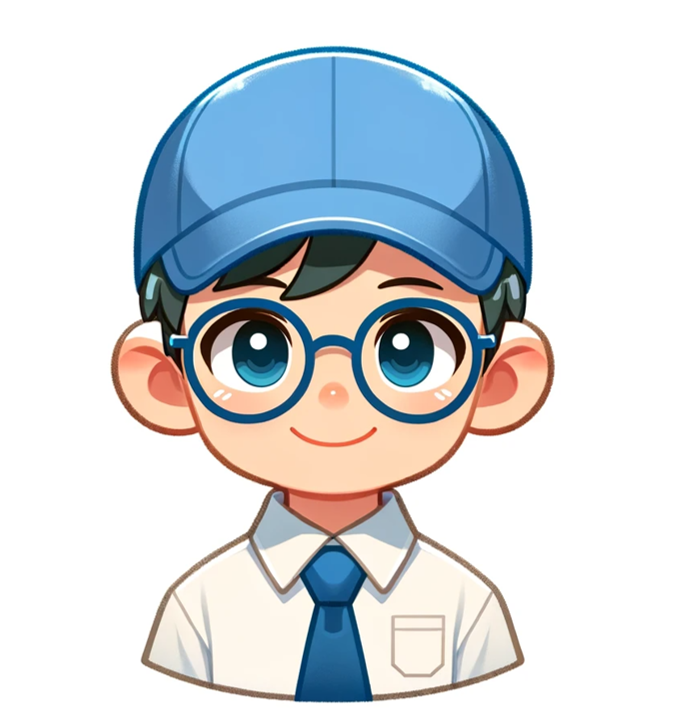}}
\end{minipage}
\textbf{Ben}: ``You are Ben, a high school student with a track record of excellent grades, particularly in mathematics. Your friends admire your diligence and often seek your guidance in their studies. Your role is to scrutinize the problem at hand with your usual attention to detail, drawing from your vast knowledge of math principles. After considering your friends' approaches, carefully construct your answer, ensuring to clarify each step of your process. Your clear and logical explanations are valuable, as they will serve as a benchmark for your friends to compare and refine their own solutions.'' \\
\vspace{-1mm}
\begin{minipage}[b]{0.16\columnwidth}
    \centering
    \raisebox{-.1in}{\includegraphics[width=\linewidth]{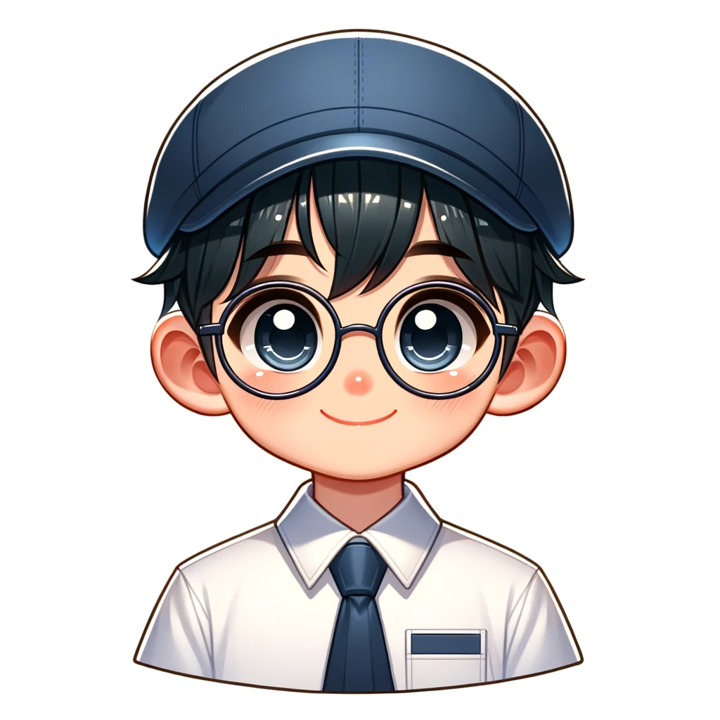}}
\end{minipage}
\textbf{Peter}: ``You are Peter, a high school student recognized for your unique problem-solving abilities. Your peers often turn to you for assistance when they encounter challenging tasks, as they appreciate your knack for devising creative solutions. Today, your challenge is to dissect the given math problem, leveraging your unique problem-solving strategies. Once you've crafted your solution, share it with your friends, Ben and Kitty, so they can see a different perspective. Your innovative approach will not only provide an answer but also inspire Ben and Kitty to think outside the box and possibly revise their own solutions.'' \\
\midrule

\underline{\textbf{\textsc{Communication Prompts}}} \\
Please consider the example provided and think it step by step. \\
Question: \{\} \\
Here is a solution process from your friend: \\
Solution: \{\} \\
Your friend's confidence in this solution is: \{\} \\
Based on your friend's solution, carefully re-examine your previous answer. If your friend's confidence level is below 0.5, it suggests a high probability that the solution might be incorrect. Remember, solutions with high confidence can also be wrong. Utilize your talent and critical thinking to provide a new step-by-step solution process. \\
\midrule
\bottomrule
\caption{Examples of system prompts used for role-playing and communication prompts in EoT communication.}
\label{table:role-prompt}
\end{tabular}
\end{table*}

\section{Case Studies}
\label{appendix:app-case_study}

To deepen our understanding of the four communication paradigms, we conducted case studies for each. The processes of these paradigms are detailed in Tables~\ref{table:case-memory}, \ref{table:case-report}, \ref{table:case-relay}, and \ref{table:case-debate}, respectively. These demonstrate that the EoT method, by introducing external insights through cross-model communication, can effectively correct reasoning errors and assist models in arriving at correct answer.

\begin{table}[ht]
\begin{tabular}{p{13cm}}

\toprule
\midrule
\textbf{Q}: Carla is downloading a 200 GB file. Normally she can download 2 GB/minute, but 40\% of the way through the download, Windows forces a restart to install updates, which takes 20 minutes. Then Carla has to restart the download from the beginning. How load does it take to download the file? \\
\midrule
\begin{minipage}[b]{0.16\columnwidth}
    \centering
    \raisebox{-.1in}{\includegraphics[width=\linewidth]{appendices/icons/Kitty.png}}
\end{minipage}
\textbf{Kitty's Solution}: Carla was able to download for 40\% of the time before the restart. 40\% of 200 GB is 80 GB. Since she can download 2 GB/minute, it would take her 80/2 = 40 minutes to download 80 GB. After the forced restart, she has to restart the download from the beginning, so she would need another 200/2 = 100 minutes to complete the download. Plus the 20 minutes for the restart, the total time would be 40 + 100 + 20 = 160 minutes. So the answer is 160 minutes.
\\
\midrule
\begin{minipage}[b]{0.16\columnwidth}
    \centering
    \raisebox{-.1in}{\includegraphics[width=\linewidth]{appendices/icons/Ben.png}}
\end{minipage}
\textbf{Ben's Solution}: Carla needs to download a 200 GB file. Normally, she can download 2 GB/minute, which means it would take her 200 / 2 = 100 minutes to download the file. However, 40\% of the way through the download, Windows forces a restart for 20 minutes, which means Carla has downloaded 200 * 0.4 = 80 GB before the restart. After the restart, she needs to start the download again from the beginning, so the total time it takes her would be 100 + (200 / 2) = 200 minutes. Therefore, it takes Carla 200 minutes to download the file. So the answer is 200 minutes.
\\
\midrule
\begin{minipage}[b]{0.16\columnwidth}
    \centering
    \raisebox{-.1in}{\includegraphics[width=\linewidth]{appendices/icons/Peter.png}}
\end{minipage}
\textbf{Peter's Response}: Upon re-evaluating the problem and reviewing Kitty and Ben's solutions, it's clear their results deviated due to a misunderstanding of the problem conditions - primarily, how much of the downloaded file was retained after Window's forced restart. To clarify, here is the new step-by-step solution: \\
1. Start with Carla's download rate: she can download 2 GB/minute. \\
2. Therefore, in theory, if uninterrupted, Carla would be able to download the 200 GB file in 200 GB / 2 GB/minute = 100 minutes. \\
3. However, 40\% of the way through the download, Windows forced a restart, so Carla had already downloaded 0.40 x 200 GB = 80 GB at that time. This took 80 GB / 2 GB/minute = 40 minutes. \\
4. The problem statement clarifies that after the restart, Carla has to start again from the beginning. This means the 80 GB downloaded before the forced restart was wasted. \\
5. Consequently, after the restart, Carla still needs to download the whole 200 GB file, which, at her rate, will take her another 200 GB / 2 GB/minute = 100 minutes. \\
6. We cannot forget the time for the Windows restart: it took an additional 20 minutes. \\
7. Consequently, the total time to download the file, taking into account the restart, is the sum of the time for the initial download attempt, the time for the restart, and the time for the full download post-restart. It results in 40 minutes (initial incomplete download) + 20 minutes (Windows restart) + 100 minutes (full download post-restart) = 160 minutes. \\
From these calculations, it appears that Kitty's solution is correct. It takes Carla 160 minutes to download the file, given the interruption due to Windows update that forced her to restart the download from the beginning. So the answer is 160 minutes.
\\
\midrule
\bottomrule
\caption{An illustrative example of the Memory paradigm of EoT in the GSM8K dataset.} 
\label{table:case-memory}
\end{tabular}
\end{table}

\clearpage
\begin{table}[ht]
\begin{tabular}{p{13cm}}

\toprule
\midrule
\textbf{Q}: Carla is downloading a 200 GB file. Normally she can download 2 GB/minute, but 40\% of the way through the download, Windows forces a restart to install updates, which takes 20 minutes. Then Carla has to restart the download from the beginning. How load does it take to download the file? \\

\midrule
\begin{minipage}[b]{0.16\columnwidth}
    \centering
    \raisebox{-.1in}{\includegraphics[width=\linewidth]{appendices/icons/Ben.png}}
\end{minipage}
\textbf{Ben's Solution}: Carla needs to download a 200 GB file. Normally, she can download 2 GB/minute, which means it would take her 200 / 2 = 100 minutes to download the file. However, 40\% of the way through the download, Windows forces a restart for 20 minutes, which means Carla has downloaded 200 * 0.4 = 80 GB before the restart. After the restart, she needs to start the download again from the beginning, so the total time it takes her would be 100 + (200 / 2) = 200 minutes. Therefore, it takes Carla 200 minutes to download the file. So the answer is 200 minutes.
\\
\midrule
\begin{minipage}[b]{0.16\columnwidth}
    \centering
    \raisebox{-.1in}{\includegraphics[width=\linewidth]{appendices/icons/Peter.png}}
\end{minipage}
\textbf{Peter's Solution}: Carla can download 2 GB in 1 minute, so to download 200 GB, it will take her 200/2 = 100 minutes if she didn't have to restart. However, she was 40\% of the way through the download when Windows forced the restart, so she had already downloaded 0.4 x 200 = 80 GB. The restart took 20 minutes, so she lost 20 x 2 = 40 GB of progress. After the restart, she had to start again from 0 GB. So in total, she had to download 200 - 80 - 40 = 80 GB. At her regular download speed of 2 GB/minute, it will take her 80/2 = 40 minutes to download the remaining 80 GB. Adding the 20 minutes of the Windows update, the total time it takes Carla to download the file is 20 + 40 = 60 minutes. So the answer is 60 minutes.
\\
\midrule

\begin{minipage}[b]{0.16\columnwidth}
    \centering
    \raisebox{-.1in}{\includegraphics[width=\linewidth]{appendices/icons/Kitty.png}}
\end{minipage}
\textbf{Kitty's Solution}: Carla was able to download for 40\% of the time before the restart. 40\% of 200 GB is 80 GB. Since she can download 2 GB/minute, it would take her 80/2 = 40 minutes to download 80 GB. After the forced restart, she has to restart the download from the beginning, so she would need another 200/2 = 100 minutes to complete the download. Plus the 20 minutes for the restart, the total time would be 40 + 100 + 20 = 160 minutes. So the answer is 160 minutes.
\\
\midrule
\begin{minipage}[b]{0.16\columnwidth}
    \centering
    \raisebox{-.1in}{\includegraphics[width=\linewidth]{appendices/icons/Kitty.png}}
\end{minipage}
\textbf{Kitty's Response}: Analyzing both Ben's solution and Peter's solution, Peter's conclusion is incorrect. He misunderstood the situation. After the Windows restart, the download does not continue from where it stopped. It starts from the beginning, which means the download progress does not reduce what is left to download. \\
Revisiting Ben's solution, he almost got everything right except the final calculation step whereas he added an extra 100 minutes. The correct calculation should be: \\
Before forced restart: 80 GB (40\% of the total file) in 40 minutes (80/2 = 40) \\
After forced restart: 200 GB total file in 100 minutes (200/2 = 100) \\
Windows restart: 20 minutes \\
Total download time = Before restart time + After restart time + Windows restart time \\
Total download time = 40 minutes + 100 minutes + 20 minutes = 160 minutes \\
Ben's setup was correct, but he added an extra 100 minutes due to an arithmetic mistake. The download time in his calculations should total to 160 minutes, not 200 minutes. So, effectively upon examining both scenarios by Peter and Ben, when taking all aspects into account and considering the restart, the total time it should take Carla to download the file is 160 minutes in total. So the answer is 160 minutes.
\\

\midrule
\bottomrule

\caption{An illustrative example of the Report paradigm of EoT in the GSM8K dataset.} 
\label{table:case-report}
\end{tabular}
\end{table}

\clearpage
\begin{table}[ht]
\begin{tabular}{p{13cm}}

\toprule
\midrule
\begin{minipage}[b]{0.16\columnwidth}
    \centering
    \raisebox{-.1in}{\includegraphics[width=\linewidth]{appendices/icons/Ben.png}}
\end{minipage}
\textbf{Ben's Solution}: Carla needs to download a 200 GB file. Normally, she can download 2 GB/minute, which means it would take her 200 / 2 = 100 minutes to download the file. However, 40\% of the way through the download, Windows forces a restart for 20 minutes, which means Carla has downloaded 200 * 0.4 = 80 GB before the restart. After the restart, she needs to start the download again from the beginning, so the total time it takes her would be 100 + (200 / 2) = 200 minutes. Therefore, it takes Carla 200 minutes to download the file. So the answer is 200 minutes. \\
\midrule
\begin{minipage}[b]{0.16\columnwidth}
    \centering
    \raisebox{-.1in}{\includegraphics[width=\linewidth]{appendices/icons/Kitty.png}}
\end{minipage}
\textbf{Kitty's Response}: After reviewing Ben's solution, it appears there seems to be a mathematical error in the final part of his solution. While he correctly calculates that Carla has downloaded 80 GB of the file before the restart, he fails to account for the time Carla already spent downloading the 80GB in his final total. Furthermore, he miscalculated the time it takes Carla to download the file after the restart. Now, let's correct the calculation. Before the restart, Carla downloaded 40\% of the file, which is 40/100 * 200 = 80 GB. Given she downloads 2 GB per minute, this would have taken her 80 / 2 = 40 minutes. Unfortunately, the computer restarts, wasting 20 minutes of her time. After the restart, Carla has to download the entire file again. With her download speed, this will take another 200 / 2 = 100 minutes. Therefore, the total time Carla spends downloading the file is the time before the restart + the time wasted on the restart + the time after the restart, which is 40 minutes + 20 minutes + 100 minutes = 160 minutes. So, the final and correct answer should indeed be 160 minutes. \\
\midrule
\begin{minipage}[b]{0.16\columnwidth}
    \centering
    \raisebox{-.1in}{\includegraphics[width=\linewidth]{appendices/icons/Peter.png}}
\end{minipage}
\textbf{Peter's Solution}: Carla can download 2 GB in 1 minute, so to download 200 GB, it will take her 200/2 = 100 minutes if she didn't have to restart. However, she was 40\% of the way through the download when Windows forced the restart, so she had already downloaded 0.4 x 200 = 80 GB. The restart took 20 minutes, so she lost 20 x 2 = 40 GB of progress. After the restart, she had to start again from 0 GB. So in total, she had to download 200 - 80 - 40 = 80 GB. At her regular download speed of 2 GB/minute, it will take her 80/2 = 40 minutes to download the remaining 80 GB. Adding the 20 minutes of the Windows update, the total time it takes Carla to download the file is 20 + 40 = 60 minutes. So the answer is 60 minutes. \\
\midrule
\begin{minipage}[b]{0.16\columnwidth}
    \centering
    \raisebox{-.1in}{\includegraphics[width=\linewidth]{appendices/icons/Ben.png}}
\end{minipage}
\textbf{Ben's Response}: Upon reviewing both solutions, it seems there might have been some misunderstandings. As per the problem, Carla can download at 2 GB/minute, and she needs to download a 200 GB file. By this rate, it would take her 200 GB / 2 GB per minute = 100 minutes without any interruptions. However, 40\% of the way through the download, Windows force restart happened which lasted for 20 minutes.  By that time, she downloaded 40\% of her file which is equal 200 GB * 0.4 = 80 GB. The restart not only consumed 20 minutes but also voided her previous download progress, which means after the updates she had to start downloading all over again. So, the total time taken by Carla to download the 200 GB file would be calculated as: Regular download time (200 GB at 2 GB/minute) + Time spent in updates = (200 / 2) + 20 = 100 + 20 = 120 minutes. So the correct total time it takes Carla to download the file is 120 minutes.\\
\midrule
\bottomrule

\caption{An illustrative example of the Relay paradigm of EoT in the GSM8K dataset. Kitty firmly maintains her correct answer, while Ben's response begins to waver.} 
\label{table:case-relay}
\end{tabular}
\end{table}

\clearpage
\begin{table}[ht]
\begin{tabular}{p{13cm}}

\toprule
\midrule
\textbf{Q}: Carla is downloading a 200 GB file. Normally she can download 2 GB/minute, but 40\% of the way through the download, Windows forces a restart to install updates, which takes 20 minutes. Then Carla has to restart the download from the beginning. How load does it take to download the file? \\
\midrule
\begin{minipage}[b]{0.16\columnwidth}
    \centering
    \raisebox{-.1in}{\includegraphics[width=\linewidth]{appendices/icons/Peter.png}}
\end{minipage}
\textbf{Peter's Response}: Analyzing both solutions, it appears that there is a slight misunderstanding in Ben's solution. He has missed considering the download progress lost due to the forced restart of Windows. Let's modify and re-explain the solution in a way that takes this into account: \\
Step 1: Calculate the time for her initial progress. Carla was able to download 40\% of the file before the interruption. That is, she downloaded 0.4 * 200GB = 80GB. With her speed of 2GB/minute, she used 80/2 = 40 minutes on this initial download. \\
Step 2: Consider the time lost due to the forced restart. The Windows forced update lasted for 20 minutes, during which time she made no progress on her download. \\
Step 3: Calculate the remaining time to finish the download. After the forced restart, Carla had to re-download the whole file. It takes her 200GB / 2GB per minute = 100 minutes at her download speed to download the whole file. \\
Step 4: Calculate the total time. The total time it takes Carla to download the whole file, considering the interruption, is the sum of the time it took for her initial progress, the time lost due to the forced restart, and the time it took for her to finish the download after the restart. This is 40 minutes + 20 minutes + 100 minutes = 160 minutes. \\
Therefore, the correct answer is 160 minutes. This improved solution accounts for the hiccup caused by the Windows update which Ben did not factor into his original problem-solving strategy.
\\
\midrule
\begin{minipage}[b]{0.16\columnwidth}
    \centering
    \raisebox{-.1in}{\includegraphics[width=\linewidth]{appendices/icons/Ben.png}}
\end{minipage}
\textbf{Ben's Response}: Reviewing the problem again, I can see that there are errors in both my solution and my friend Peter's solution. Here is the corrected step-by-step solution: \\
1. Carla downloaded 40\% of the 200 GB file before the system restart. This equals 40/100 * 200 = 80 GB. \\
2. At a rate of 2 GB/minute, it must have taken her 80 / 2 = 40 minutes to download that 80 GB. \\
3. Then, the Windows update interrupted the download for 20 minutes. \\
4. After the restart, she had to start again from the beginning, so she needs again 200 GB/2 GB per minute = 100 minutes to download the full file. \\
5. If we add up these times, we get the total time Carla spent on the download: 40 + 20 + 100 = 160 minutes. \\
So, it would take Carla 160 minutes to download the file, correcting both my previous response and Peter's calculations. This solution process allows for the clear understanding of the time taken to download the file considering both the time before and after the interruption. So the answer is 160 minutes. \\
\midrule
\bottomrule
\caption{An illustrative example of the Debate paradigm of EoT in the GSM8K dataset. Through mutual discussion, Ben and Peter identify the flaws in their respective solutions and arrive at the correct answer.} 
\label{table:case-debate}
\end{tabular}
\end{table}

\end{document}